\documentclass[sigconf]{acmart}

\usepackage{hyperref}
\usepackage{graphicx}
\usepackage{subfig}
\usepackage{float}
\usepackage{algorithm}
\usepackage{multirow}
\usepackage{bm}
\usepackage{amsmath}
\usepackage{algorithmic}

\def\vtheta{{\bm{\theta}}}

\def\vc{{\bm{c}}}

\def\vm{{\bm{m}}}
\def\vn{{\bm{n}}}

\def\vp{{\bm{p}}}
\def\vq{{\bm{q}}}

\def\vx{{\bm{x}}}

\def\vz{{\bm{z}}}

\let\sss = \scriptscriptstyle

\newcommand{\R}{\mathbb{R}}

\AtBeginDocument{%
  }

\copyrightyear{2024} 
\acmYear{2024} 
\setcopyright{acmlicensed}\acmConference[KDD '24]{Proceedings of the 30th ACM SIGKDD Conference on Knowledge Discovery and Data Mining}{August 25--29, 2024}{Barcelona, Spain}
\acmBooktitle{Proceedings of the 30th ACM SIGKDD Conference on Knowledge Discovery and Data Mining (KDD '24), August 25--29, 2024, Barcelona, Spain}
\acmDOI{10.1145/3637528.3671887}
\acmISBN{979-8-4007-0490-1/24/08}

\begin{document}

\title{URRL-IMVC: Unified and Robust Representation Learning for Incomplete Multi-View Clustering}

\author{Ge Teng}
\authornote{This work was done during a research internship at Alibaba Cloud.}
\email{12115044@zju.edu.cn}
\orcid{0000-0002-1331-9868}
\affiliation{%
  \institution{Zhejiang University}
  \city{Hangzhou}
  \country{China}
}

\author{Ting Mao}
\email{maoting.mao@alibaba-inc.com}
\orcid{0009-0001-9531-6328}
\affiliation{%
  \institution{Alibaba Cloud}
  \city{Hangzhou}
  \country{China}
}

\author{Chen Shen}
\authornote{Corresponding Author}
\email{jason.sc@alibaba-inc.com}
\orcid{0000-0002-7534-0830}
\affiliation{%
  \institution{Alibaba Cloud}
  \city{Hangzhou}
  \country{China}
}

\author{Xiang Tian}
\authornotemark[2]
\email{tianx@zju.edu.cn}
\orcid{0000-0003-0735-8454}
\affiliation{%
  \institution{Zhejiang University}
  \city{Hangzhou}
  \country{China}
}
\affiliation{%
  \institution{Zhejiang University Embedded System Engineering Research Center, Ministry of Education of China}
  \city{Hangzhou}
  \country{China}
}

\author{Xuesong Liu}
\email{11015006@zju.edu.cn}
\orcid{0000-0001-8549-0368}
\affiliation{%
  \institution{Zhejiang University}
  \city{Hangzhou}
  \country{China}
}
\affiliation{%
  \institution{Zhejiang University Embedded System Engineering Research Center, Ministry of Education of China}
  \city{Hangzhou}
  \country{China}
}

\author{Yaowu Chen}
\email{yaowuchen@zju.edu.cn}
\orcid{0000-0001-7266-1535}
\affiliation{%
  \institution{Zhejiang University}
  \city{Hangzhou}
  \country{China}
}
\affiliation{%
  \institution{Zhejiang University Embedded System Engineering Research Center, Ministry of Education of China}
  \city{Hangzhou}
  \country{China}
}

\author{Jieping Ye}
\email{yejieping.ye@alibaba-inc.com}
\orcid{0000-0001-8662-5818}
\affiliation{%
  \institution{Alibaba Cloud}
  \city{Hangzhou}
  \country{China}
}

\renewcommand{\shortauthors}{Ge Teng et al.}

\begin{abstract}
  Incomplete multi-view clustering (IMVC) aims to cluster multi-view data that are only partially available. This poses two main challenges: effectively leveraging multi-view information and mitigating the impact of missing views. Prevailing solutions employ cross-view contrastive learning and missing view recovery techniques. However, they either neglect valuable complementary information by focusing only on consensus between views or provide unreliable recovered views due to the absence of supervision. To address these limitations, we propose a novel Unified and Robust Representation Learning for Incomplete Multi-View Clustering (URRL-IMVC). URRL-IMVC directly learns a unified embedding that is robust to view missing conditions by integrating information from multiple views and neighboring samples. Firstly, to overcome the limitations of cross-view contrastive learning, URRL-IMVC incorporates an attention-based auto-encoder framework to fuse multi-view information and generate unified embeddings. Secondly, URRL-IMVC directly enhances the robustness of the unified embedding against view-missing conditions through KNN imputation and data augmentation techniques, eliminating the need for explicit missing view recovery. Finally, incremental improvements are introduced to further enhance the overall performance, such as the Clustering Module and the customization of the Encoder. We extensively evaluate the proposed URRL-IMVC framework on various benchmark datasets, demonstrating its state-of-the-art performance. Furthermore, comprehensive ablation studies are performed to validate the effectiveness of our design.
\end{abstract}





\maketitle

\section{Introduction}
\label{section: Introduction}
Multi-view data \citep{MVC-Survey-2} is commonly collected and utilized in various domains, making multi-view clustering (MVC) a crucial tool for analyzing such data and uncovering its underlying structures \citep{MVC-Survey, MVC-Survey-3}. Previous research has proposed several approaches \citep{MFLVC, Multi-VAE} achieving promising performance by exploiting consensus or complementary information between views. However, in real-world applications, some views may be partially unavailable due to sensor malfunctions or other practical reasons. Existing MVC methods heavily rely on complete views to learn a comprehensive representation for clustering, making them inadequate under such conditions. To address this issue, Incomplete Multi-view Clustering (IMVC) methods have been introduced to reduce the impact of missing views \citep{IMVC-Survey}. Various IMVC approaches have been proposed, including matrix decomposition \citep{PMVC}, kernel-based \citep{IMKKC}, and graph-based \citep{SCIMC} methods. With the superior feature representation ability demonstrated by deep learning, some IMVC methods have integrated deep learning techniques, known as Deep Incomplete Multi-view Clustering (DIMVC) methods, which we will mainly discuss below. The key challenges in the IMVC task revolve around two problems: i) effectively utilizing multi-view information, and ii) mitigating the impact of missing views. Previous DIMVC works \citep{PMVC-CG, Completer, DCP, CPSPAN, RecFormer} have employed two mainstream strategies to address these problems: 1) cross-view contrastive learning, and 2) missing view recovery. However, these strategies have inherent drawbacks.

\begin{figure*}[htbp]
    \centering
    \subfloat[Cross-view contrastive learning framework]{
        \includegraphics[width=0.4\linewidth]{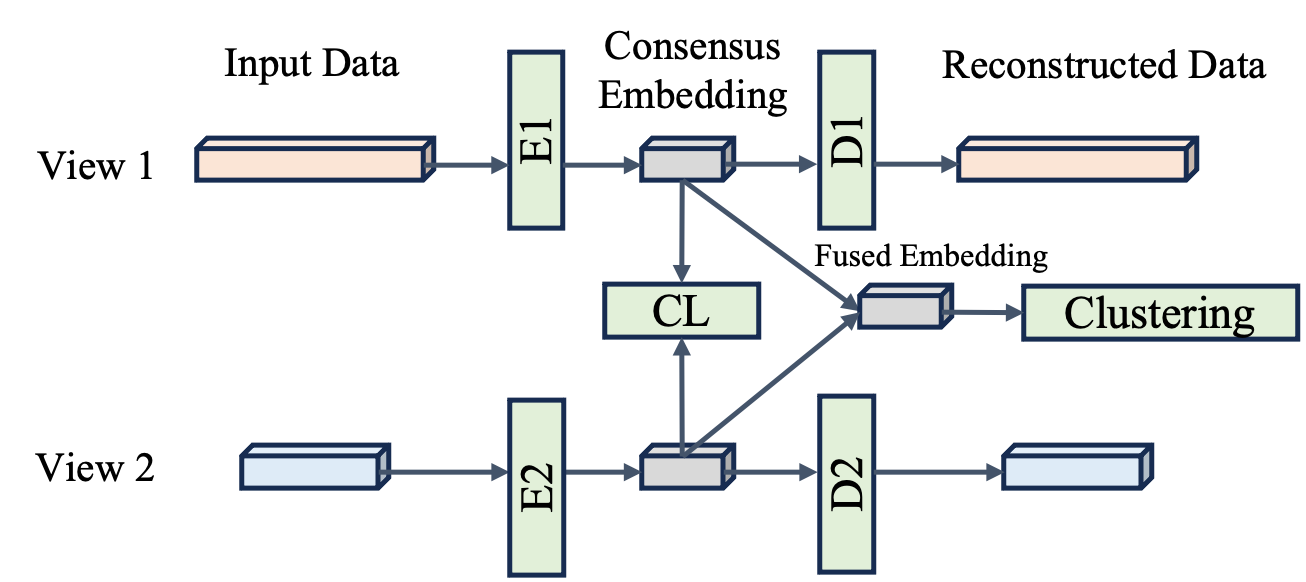}
        \label{fig: pipeline 1}
    }
    \subfloat[Missing view recovery framework]{
        \includegraphics[width=0.5\linewidth]{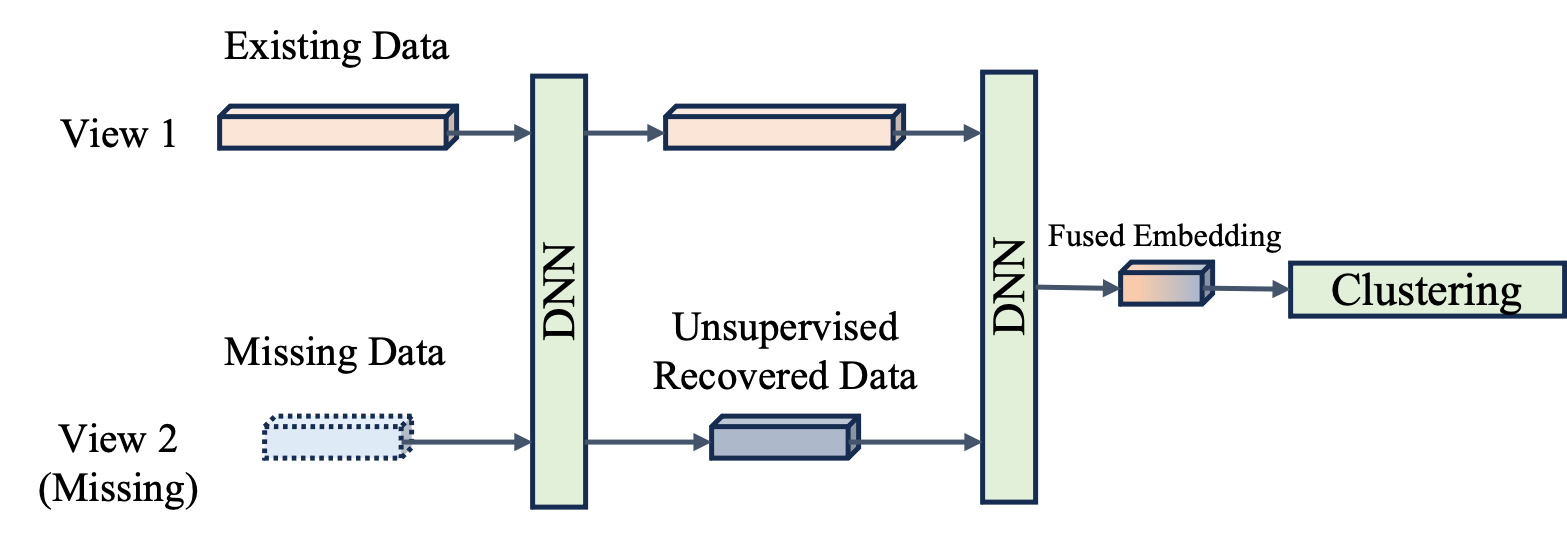}
        \label{fig: pipeline 2}
    } \\
    \subfloat[Our unified and robust learning framework]{
        \includegraphics[width=0.6\linewidth]{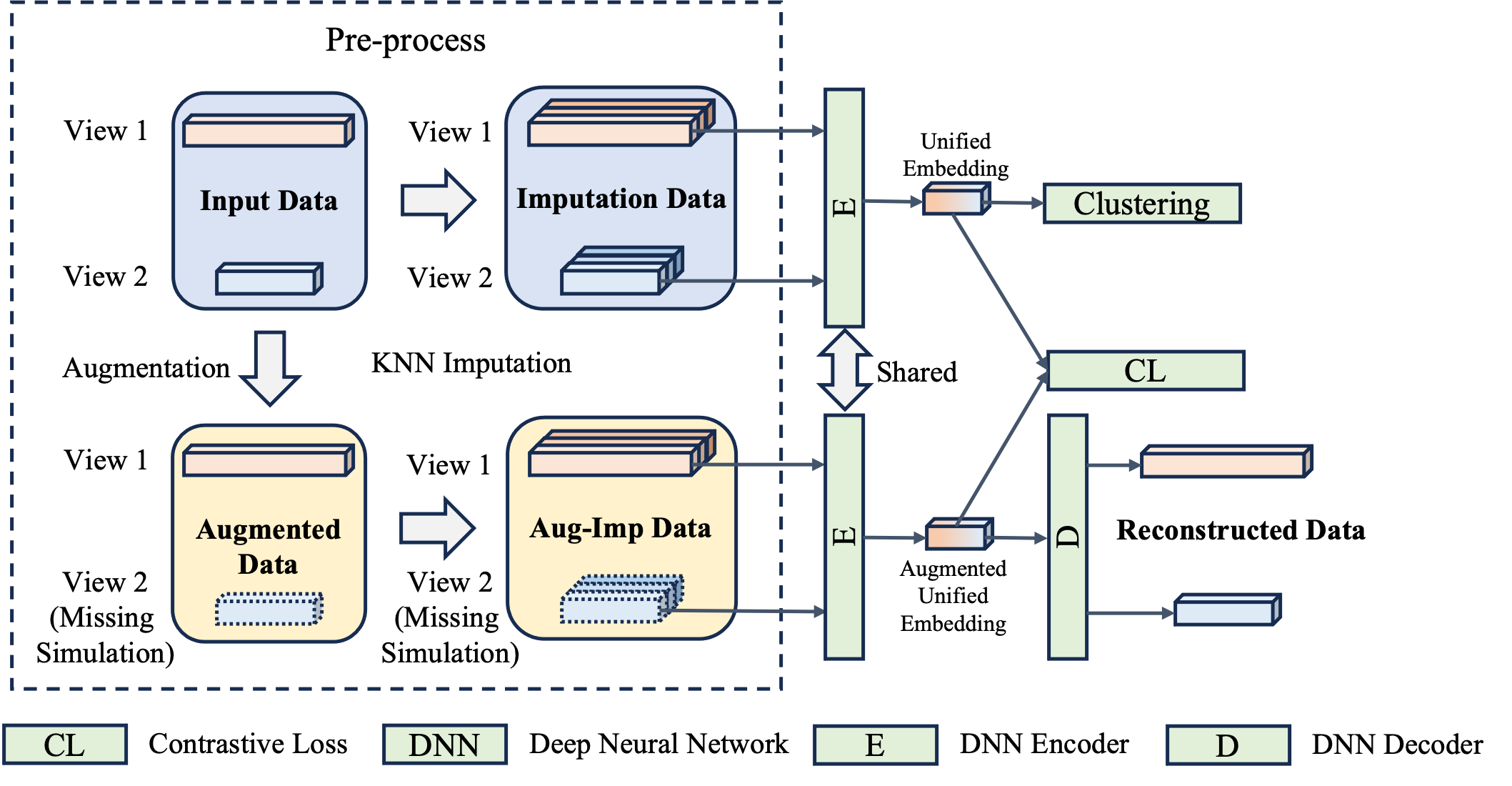}
        \label{fig: pipeline 3}
    }
    \caption{A comparison between our learning framework and commonly used cross-view contrastive learning and missing view recovery framework. The key difference lies in how the unified embedding for clustering is obtained. Our design (\ref{fig: pipeline 3}) directly fuses multi-view information and utilizes KNN imputation and data augmentation to obtain unified and robust embedding under view-missing conditions, avoiding the drawbacks of (\ref{fig: pipeline 1}) and (\ref{fig: pipeline 2}).}
    \Description{Our framework avoids the drawbacks of cross-view contrastive learning and missing view recovery, by directly learning a unified and robust embedding.}
    \label{fig: pipeline}
\end{figure*}

A general framework for cross-view contrastive learning is illustrated in Fig \ref{fig: pipeline 1}, which originates from the MVC approaches. In this framework, Deep Neural Network (DNN) auto-encoders are employed to extract embeddings for each view. The embeddings are then aligned using a contrastive loss, aiming to minimize the distance between embeddings from the same sample across different views while simultaneously maximizing the distance with other samples \citep{CPSPAN, Completer, SURE}. However, this framework primarily focuses on extracting consensus information in multi-view data, overlooking the valuable complementary information present. Additionally, the efficiency of the pair-wise contrastive strategy suffers as the number of views increases, and the effectiveness of this strategy diminishes due to less overlapped information between views (See Table \ref{table: different views} for experimental analysis). Theoretical analysis by \citet{ESSCA} supports these observations, highlighting that contrastive alignment can reduce the number of separable clusters in the representation space, with this effect worsening as the number of views increases.

The missing view recovery framework, as depicted in \ref{fig: pipeline 2}, is commonly adopted in IMVC approaches. Typically, a DNN is employed to recover the missing view, either in the data or latent space. Subsequently, MVC methods or another view fusion network are utilized for clustering based on the recovered views. However, the reliability of the recovered views is a concern since the recovery ability of DNNs relies on unsupervised training. Meanwhile in some instances, \citep{RecFormer} for example, missing views are recovered by a fused embedding in the first stage, and subsequently used to generate another fused embedding for clustering in the second stage, introducing unnecessary complexity and inefficiency to the pipeline. We propose that a well-designed recovery-free method can achieve comparable performance to recovery-based methods while offering the advantages of simplicity and reduced computational overhead.

To address the aforementioned challenges, we propose a Unified and Robust Representation Learning framework for Incomplete Multi-View Clustering (URRL-IMVC). Our framework, depicted in Fig \ref{fig: pipeline 3}, is designed to be cross-view contrastive learning-free and missing view recovery-free. First, to overcome the limitations of cross-view contrastive learning, we propose a new framework that fuses multi-view information into a unified embedding instead of contrasting each view's information. 
We achieve this by designing an attention-based auto-encoder network, which captures both consensus and complementary information and intelligently fuses them. Moreover, it is naturally scalable to different numbers of views.
Second, to tackle the issue of missing views, we aim to directly enhance the robustness of the unified embedding against view-missing conditions without explicitly recovering the missing views. We introduce two strategies to achieve this robustness. 1) We treat view missing as a form of noise and draw inspiration from successful applications of denoising and masked auto-encoders \citep{DAE, MAE}. Our proposed approach randomly drops out existing views as a form of data augmentation to simulate the view missing condition. By reconstructing denoised input data from the unified embedding and imposing constraints between the augmented and un-augmented embeddings, we enhance the robustness of the unified representation. 2) As the old saying goes, ``One cannot make bricks without straw'', it is hard to learn to reconstruct a dropped-out view directly. We introduce k-nearest neighbors (KNN) as additional inputs, with a cross-view imputation strategy to fill in the missing or dropped-out views, providing valuable hints for reconstruction. We want to highlight that while previous methods have focused on either fusing multi-view information \citep{GP-MVC, CMVHHC} or incorporating neighborhood information \citep{ClusFormer, L-GCN, VE-GCN, DFCN} for clustering, our approach represents one of the initial endeavors to fuse both aspects.
Finally, we conduct experiments based on this framework and make incremental improvements to enhance clustering performance and stability. Some of the key enhancements include the customization of the Transformer-based Encoder to filter out noise and emphasize critical information, and the introduction of the Clustering Module to learn clustering-friendly representations.

To summarize, our main contributions are:

\begin{itemize}
\item \textbf{Unified}: We propose a unified representation learning framework that efficiently fuses both multi-view and neighborhood information, allowing for better capturing of consensus and complementary information while avoiding the limitations of cross-view contrastive learning.
\item \textbf{Robust}: We proposed novel strategies, including KNN imputation and data augmentation, to directly learn a robust representation capable of handling view-missing conditions without explicit missing view recovery.
\item \textbf{Improvements}: Multiple incremental improvements are introduced for better clustering performance and stability, including the extra Clustering Module and the customization of the Transformer-based Encoder.
\item \textbf{Experiments}: Through comprehensive experiments on diverse benchmark datasets, we demonstrate the state-of-the-art performance of our unified representation learning framework. Thorough ablation studies are also conducted to provide valuable insights for future research in this field.
\end{itemize}

\section{Related works}

Deep neural networks (DNNs) have shown good performance in learning feature representation, which is beneficial for the IMVC task. Various IMVC approaches have integrated DNNs into their framework, denoted as DIMVC approaches. In terms of network architecture, DIMVC approaches can be divided into four categories. (1) Auto-encoder-based approaches \citep{CMVHHC, CPSPAN, Completer, DCP}. These approaches utilize auto-encoders to extract high-level features of each view, which are usually combined with contrastive learning or cross-view prediction to handle the incompleteness problem. (2) Generative network-based approaches. For the IMVC task, an intuitive solution is to complete the missing views with generative models, transforming it into an MVC task. Adversarial learning \citep{GAN} is commonly adopted by generative IMVC approaches including AIMVC \citet{AIMVC}, PMVC-CG \citet{PMVC-CG}, and GP-MVC \citet{GP-MVC} to improve data distribution learning in the context of IMVC. (3) Graph Neural Network-based (GNN-based) approaches \citep{AGCL, PMVC-CG}. These approaches aim to learn consensus representations from the structure information contained in the graphs constructed for each view. (4) Transformer \citep{Transformer} or attention-based approaches. The Transformer network has gained attention in recent years due to its successful application in various domains. Its architecture, along with its Multi-head Attention mechanism, has been particularly effective in capturing complex relationships. In the field of DIMVC, RecFormer \citep{RecFormer} proposed a Transformer auto-encoder with a mask to recover missing views, while MCAC \citep{MCAC} and IMVC-PBI \citep{IMVC-PBI} incorporated attention mechanisms into their frameworks. In this paper, we leverage an auto-encoder architecture based on the Transformer framework to address the challenges of the IMVC task.

\section{The Proposed Method}

\textbf{Notations.} 
An incomplete multi-view dataset with $N$ samples and $V$ views is denoted as  $X=\{X^{(1)},X^{(2)},\cdots,X^{(V)}\}, X^{(v)}\in \R^{N\times d_v}$, where $d_v$ denotes the dimension of $v$-th view. The view missing condition can be described by a binary missing indicator matrix $M \in \{0,1\}^{N \times V}$, where $M_{ij}=0$ indicates the $j$-th view of the $i$-th sample is missing and $M_{ij}=1$ just the opposite. An extra restriction is imposed:  $\sum_jM_{ij} \geq 1$,  ensuring that at least one view is available for each sample, which is essential for the clustering task. 

\subsection{Framework}
\begin{figure*}[htbp]
    \centering
    \includegraphics[width=\linewidth]{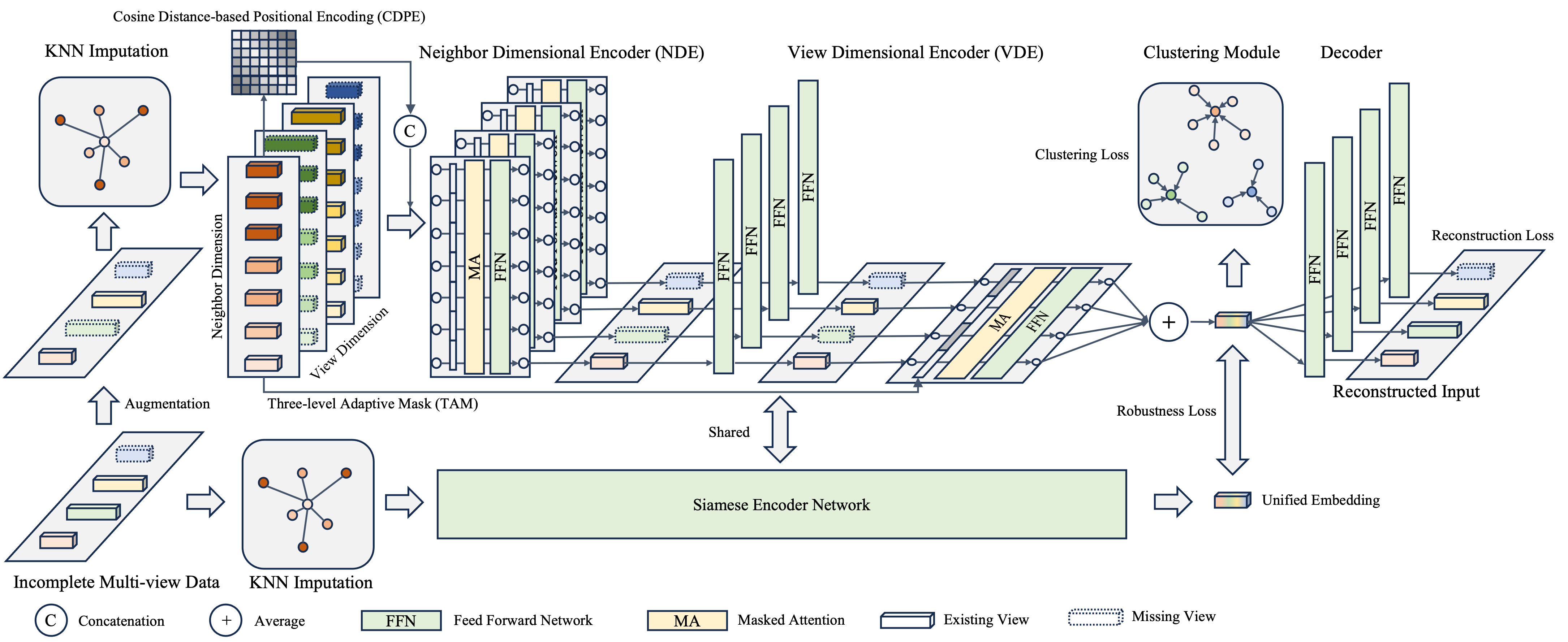}
    \caption{The overall architecture of URRL-IMVC. During training, the input data is augmented to simulate view-missing conditions, and KNN Imputation provides hints for missing views, forming an input batch with both neighbor and view dimensions. This batch is fed into the auto-encoder network, consisting of the Encoder (including the Neighbor Dimensional Encoder and View Dimensional Encoder), the Decoder, and the Clustering Module. The Encoders fuse information from the neighbor and view dimensions to generate a unified embedding. The Decoder reconstructs the augmented input, and the Clustering Module produces clustering results. Additionally, an un-augmented embedding is obtained by passing the original input data through the shared Encoders. Three loss functions, including Reconstruction loss, Robustness loss, and Clustering loss, enhance robustness against view-missing conditions and encourage learning clustering-friendly embeddings. }
    \Description{The architecture can be better understood together with the equations below.}
    \label{fig: network architecture}
\end{figure*}

Unlike many prior approaches in the field of MVC that employ view-specific auto-encoders for each view, we propose a novel framework using a unified auto-encoder that effectively fuses multi-view data. The network architecture, depicted in Fig \ref{fig: network architecture}, consists of three key modules: the Encoder $f$, the Decoder $g$, and the Clustering Module $h$. To provide a formal description, the framework operates as follows. Given an incomplete multi-view data sample $\vx = \{\vx^{(1)},\vx^{(2)},\cdots,\vx^{(V)}\}, \vx^{(v)} \in \R^{d_v}$ from dataset $X$ with its missing indicator vector $\vm \in \{0,1\}^{V}$, we apply KNN Imputation and Data Augmentation (KIDA), as described in section \ref{subsec: DAI}, to obtain the input for the auto-encoder network, 
\begin{equation}
\label{equation: KIDA}
    \begin{split}
        &\bar\vx, \bar\vx', \bar\vm, \bar\vm' = KIDA(\vx, \vm, X, M) \\ &\bar{\vx}^{(v)}, \bar{\vx}'^{(v)} \in \R^{k \times d_v}; \ \bar{\vm}, \bar{\vm}' \in \{0,1\}^{k \times V}
    \end{split}
\end{equation}
where $\bar\vx$, $\bar\vm$ is the data and mask after KNN Imputation, while $\bar\vx'$, $\bar\vm'$ is the augmented version of $\bar\vx$, $\bar\vm$, and $k$ is the hyperparameter $k$ in KNN. Note that though KNN Imputation is widely applied in prior IMVC works, it is mainly used for recovering missing views, which is different from our usage as a pre-process. Next, these inputs are fed into the Encoder network to obtain the augmented and un-augmented embeddings, denoted as $\vz'$ and $\vz$ respectively, 
\begin{equation}
\label{equation: Encoder}
    \vz = f(\bar{\vx},\bar{\vm}; \vtheta_E), \ 
    \vz' = f(\bar{\vx}',\bar{\vm}'; \vtheta_E);\ \vz, \vz' \in \R^{d_e}   
\end{equation}
where $\vtheta_E$ represents the Encoder's parameters and $d_e$ is the dimension of the embedding. Then, the Decoder maps the augmented embedding back to the data space to reconstruct the data sample, 
\begin{equation}
\label{equation: Decoder}
    \hat{\vx}' = g(\vz'; \vtheta_D), \ \hat{\vx}'^{(v)} \in \R^{d_v}
\end{equation}
where $\vtheta_D$ represents the parameters of the Decoder. Simultaneously, clustering is performed using the un-augmented embedding, 
\begin{equation}
\label{equation: cluster}
    \vc = h(\vz;\vtheta_C),\ \vc \in [0,1]^{d_c}    
\end{equation}
where $\vc$ is the clustering result, and represents the probabilities of the data sample belonging to $d_c$ cluster centers. During training, the loss function defined in equation \ref{equation: Loss} is computed to optimize parameters $\vtheta_E, \vtheta_D, \vtheta_C$; During testing, $\vc$ is regarded as the final clustering result.

In the following sections, we will introduce the Encoder module, including its two submodules: the Neighbor Dimensional Encoder (NDE) and the View Dimensional Encoder (VDE), the Decoder module, and the Clustering Module respectively.

\subsubsection{Neighbor Dimensional Encoder}
\label{subsubsec:NDE}
KNN Imputation (Section \ref{subsec: DAI}) provides additional information for missing views, but the retrieved nearest neighbors may contain noise and be unreliable. To address this issue, we propose the Neighbor Dimensional Encoder (NDE),  which is a series of customized Transformer Encoders \citep{Transformer}, with each one dedicated to a view to fuse its KNN input and filter out noise, formulated as:
\begin{equation}
    \vx_{\sss NDE} = \{\vx_{\sss NDE}^{(1)},\vx_{\sss NDE}^{(2)},\cdots,\vx_{\sss NDE}^{(V)}\},\ \vx_{\sss NDE}^{(v)} \in \R^{d_v}
\end{equation}
\begin{equation}
\label{equation: NDE2}
    \vx_{\sss NDE}^{(v)} = f_{\sss NDE}^{(v)}(CDPE(\bar{\vx}^{(v)}), \bar\vm; \vtheta_{\sss NDE}^{(v)})_{[0,:]}
\end{equation}

The $v$th Transformer Encoder corresponding to the $v$th view is represented with $f_{\sss NDE}^{(v)}(\cdot \ ;\vtheta_{\sss NDE}^{(v)})$ in equation \ref{equation: NDE2}, and $\vtheta_{\sss NDE}^{(v)}$ is its parameters. The input KNN sequence from the $v$th view $\bar{\vx}^{(v)}$ is first processed to add the Cosine Distance-based Positional Encoding (CDPE), then it is passed through the Transformer Encoder with the KNN mask $\bar\vm$. Finally, only the first vector from the output sequence is chosen as the output, denoted as $[0,:]$.

Below we introduce the two key customizations of the Transformer Encoders in NDE: the CDPE and the output choice.

\paragraph{Cosine Distance-based Positional Encoding (CDPE)}
The order or distance of the KNN instances contains vital information regarding the reliability of the inputs, with farther neighbors noisier and less reliable. To capture this information for the permutation invariant Transformer structure, we introduce Positional Encoding (PE) to provide this extra KNN order information. We explored various positional encoding (PE) designs considering the data sources and their combination with data. Among these configurations, concatenating cosine distance-based (inspired by \citet{ClusFormer}) or learnable PE with the input yielded the best results. For better interpretability, Cosine Distance-based Positional Encoding (CDPE) is chosen as our final design.
The CDPE can be explained as, 
\begin{equation}
    CDPE(\bar\vx^{(v)}) = \bar{\vx}^{(v)} \oplus d(\bar{\vx}^{(v)}), \ d(\bar{\vx}^{(v)}) \in \R^{k \times k}
\end{equation}
in which $d()$ is the function calculating the pair-wise distance of $k$ vectors and return a $k \times k$ distance matrix, and $\oplus$ stands for matrix concatenation. Given two input vectors $\vx_1$ and $\vx_2$, the pair-wise cosine distance is formulated as,
\begin{equation}
    d_{cos}(\vx_1,\vx_2) = 1 - \frac{\vx_1 \cdot \vx_2}{||\vx_1|| \cdot ||\vx_2||}
\end{equation}
We conjecture that CDPE is the most suitable for two reasons. First, it contains the KNN distance information rather than simply providing order information. Second, its value range is 0-2, which is more stable than other distance functions, e.g., Euclidean distance.

\paragraph{Output choice} (Figure \ref{fig: output choice 1})
Generally, for fusing information with a Transformer Encoder, an additional token like [CLS] can be added \cite{BERT}. However, in our unsupervised task, adding such a meaningless token can introduce noise and lead to performance degradation. Instead, we adopt the first vector of the output sequence. This design not only avoids extra noise but also introduces a bias on the first input. The first input is always the most reliable sample, i.e., the center sample for an existing view or the nearest neighbor for a missing view. By introducing this bias, important information is emphasized while fusing KNN information.

\subsubsection{View Dimensional Encoder}
\label{subsubsec:VDE}
The View Dimensional Encoder (VDE) is designed to fuse view representations and obtain unified embedding. As depicted in Figure \ref{fig: network architecture}, it consists of two parts, with firstly a Feed-Forward Network (FFN) to map the representations of different dimensions to the same latent space, and then followed by a Transformer Encoder for fusion. The FFN consists of three fully connected (FC) layers, without normalization or dropout layers, which can be detrimental to the stability of training. The FFN of VDE can be formulated as:
\begin{equation}
    \vx_{\sss VDE-F} = \vx_{\sss VDE-F}^{(1)} \oplus \vx_{\sss VDE-F}^{(2)} \oplus \cdots \oplus \vx_{\sss VDE-F}^{(V)}, \ \vx_{\sss VDE-F} \in \R^{V \times d_e}
\end{equation}
\begin{equation}
    \vx_{\sss VDE-F}^{(v)} = \sigma(\sigma (\vx_{\sss NDE}^{(v)} \bm{W}_1^{(v)} + \bm{b}_1^{(v)}) \bm{W}_2^{(v)} + \bm{b}_2^{(v)}) \bm{W}_3^{(v)} + \bm{b}_3^{(v)}
\end{equation}
In the equation, $\oplus$ represents the concatenate operation, $\sigma$ is the activation function, $\bm{W}$ and $\bm{b}$ are the weight matrix and bias vector of the FC layer respectively.

The Transformer Encoder part of the VDE can be explained as, 
\begin{equation}
    \vz = \sum_{v=1}^{V} f_{\sss VDE-T}(\vx_{\sss VDE-F}, TAM(\bar\vm, \vm); \vtheta_{\sss VDE-T}) / V
\end{equation}
in which $f_{\sss VDE-T}(\cdot \ ; \vtheta_{\sss VDE-T})$ is the VDE Transformer Encoder structure. The view representations $\vx_{\sss VDE-F}$ are passed through the Transformer Encoder along with the generated Three-level Adaptive Mask (TAM) $TAM(\bar\vm, \vm)$. The output sequence is averaged for fusion. The Transformer Encoder in VDE is also customized but different from that in NDE. The key difference lies in that views are permutation invariant, i.e., changing the order of views should yield the same output, while KNN has an order. Based on this difference, the two key customizations of the VDE Transformer are designed as:

\paragraph{Three-level Adaptive Masking (TAM)}
We employ a masking mechanism in VDE to emphasize the reliability of the inputs, instead of the positional encoding used in NDE to maintain permutation invariant. 
The self-attention in Transformer is formulated as, 
\begin{equation}
    Attention(Q,K,V,M_A) = softmax(\frac{QK^T}{\sqrt{d_k}}+M_A)V
\end{equation}
where $M_A$ is the mask applied, with negative infinity for masking and 0 for not. The input view representations can be roughly divided into 3 categories based on data completeness: (1) complete, (2) missing view with KNN imputation, and (3) missing view without imputation. Therefore, instead of the original binary mask, we design a Three-level Adaptive Mask (TAM) for the 3 categories, the mask values range from completely unmasked (1) to fully masked (3), with an intermediate level (2) in between, formulated as,
\begin{equation}
\label{equation: TLAM}
    M_A = TAM(\bar\vm, \vm) = \left\{
    \begin{aligned}
        0, & \ \vm_j = 1 \\
        \gamma, & \ \sum_{i=1}^{k}\bar\vm_{ij}>0 \ \&\  \vm_j = 0 \\
        -\infty, & \ \sum_{i=1}^{k}\bar\vm_{ij}=0 \ \&\  \vm_j = 0 \\
    \end{aligned}
    \right.
\end{equation}
in which $\vm$ and $\bar\vm$ are the original and KNN imputation generated missing matrix respectively. $\gamma$ is a negative hyperparameter to control the emphasizing intensity.

\begin{figure}[htbp]
    \centering
    \subfloat[Neighbor Dimensional Encoder]{
        \includegraphics[width=0.9\linewidth]{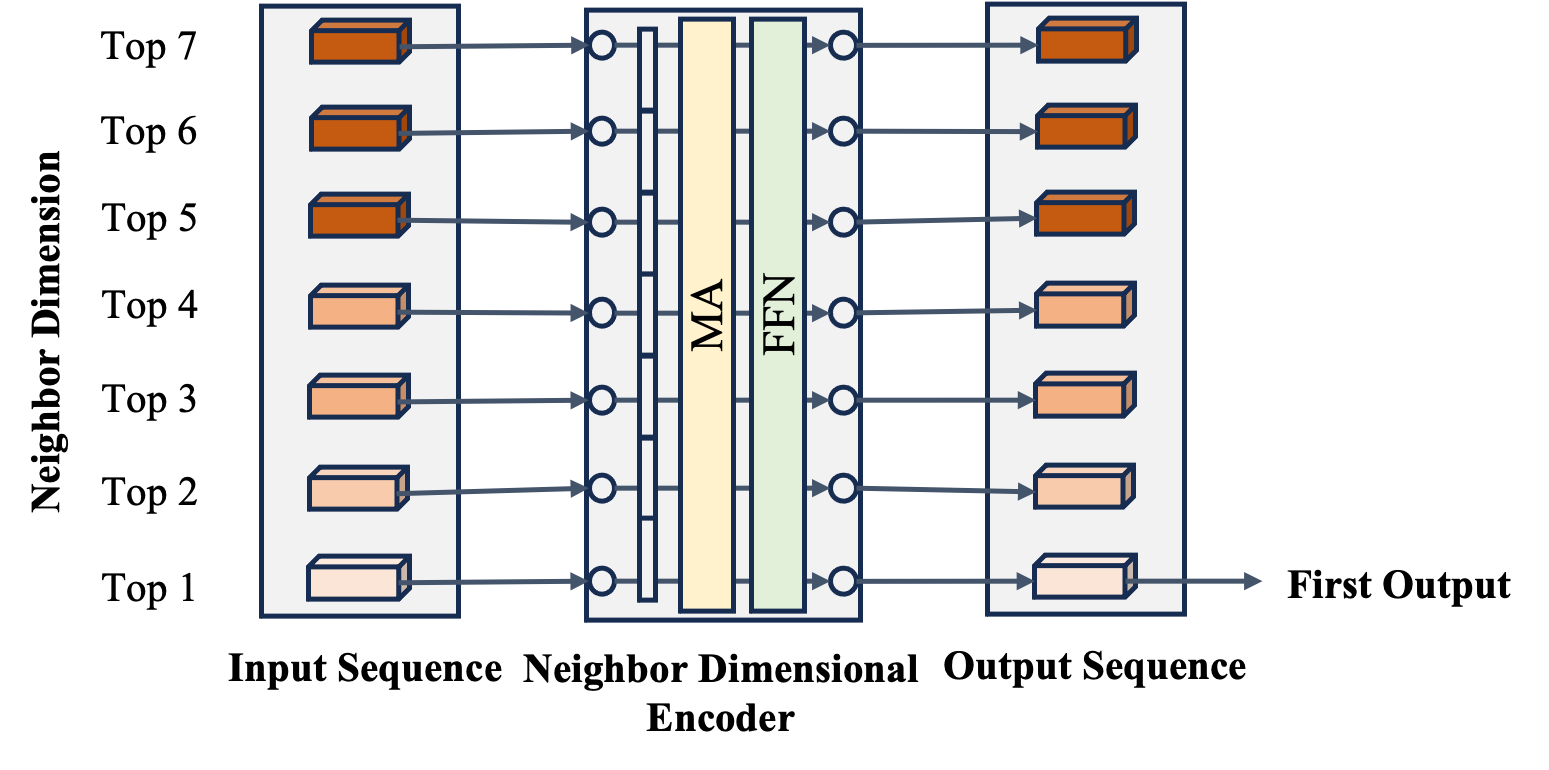}
        \label{fig: output choice 1}
    } \\ 
    \subfloat[View Dimensional Encoder]{
        \includegraphics[width=1.0\linewidth]{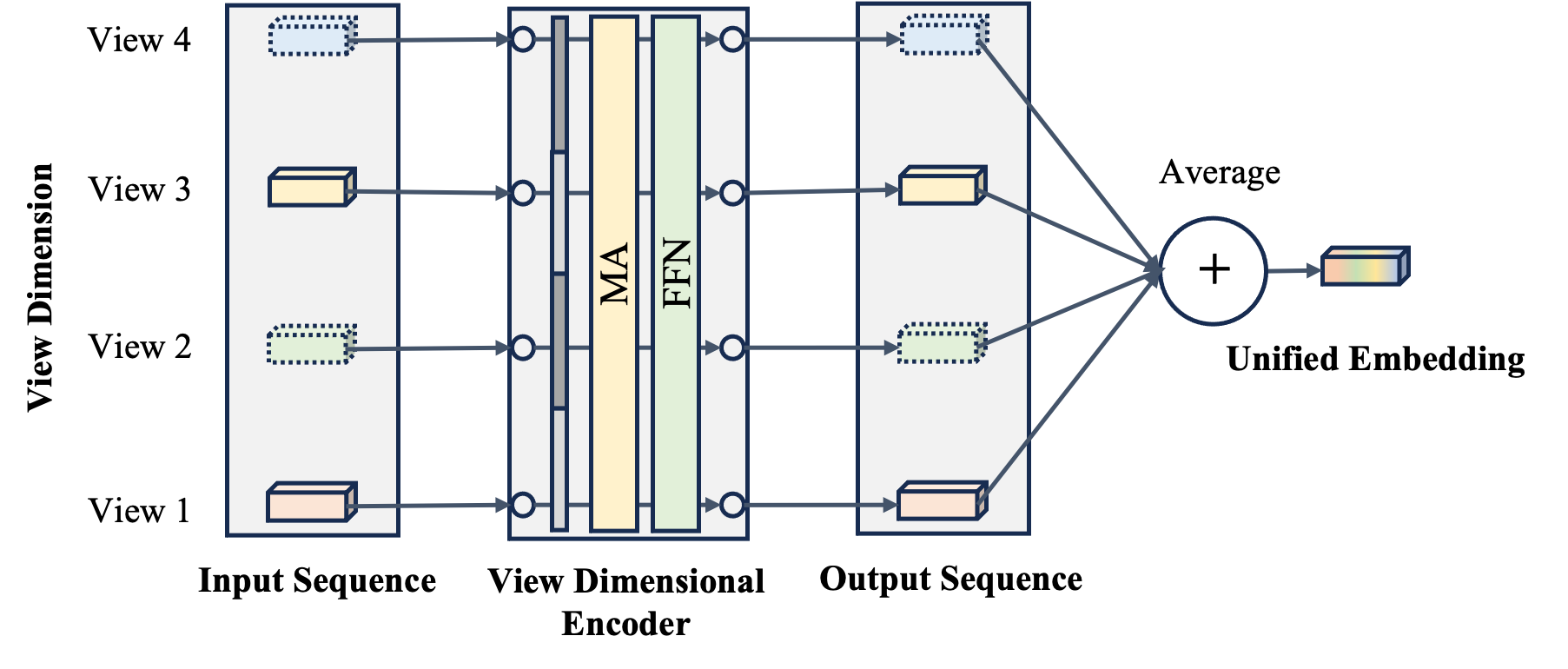}
        \label{fig: output choice 2}
    }
    \caption{An intuitive visualization of the output choice of the Neighbor Dimensional Encoder (NDE) and View Dimensional Encoder (VDE). In NDE, the first vector of the output sequence is chosen to provide a bias on the most reliable input. In VDE, the outputs are averaged to provide an unbiased representation of all views.}
    \Description{The effectiveness of this design is proved by ablation studies in the Appendix.}
    \label{fig: output choice}
\end{figure}

\paragraph{Output choice} (Figure \ref{fig: output choice 2})
To ensure permutation invariance and avoid bias towards any views, the embedding is generated by averaging all Transformer output vectors. The output choices of NDE and VDE are depicted in Figure \ref{fig: output choice} for intuitive understanding.

\subsubsection{Decoder}
The Decoder in our model is designed as a compact 4-layer FFN, to reconstruct the input from the unified embedding. Similar to the FFN in VDE, we removed its normalization and dropout layer for better stability. Through our experiments, we have observed that a deep and complex Decoder does not necessarily improve the clustering performance and may even have negative effects in certain cases. One possible explanation for this phenomenon is that a shallow and simple Decoder serves as a regularization technique on the embedding space, and prevents it from collapsing. 
This regularization effect is similar to the Locality-preserving Constraint proposed by \citet{DEN}, which helps preserve the local structure of the data. The process of Decoder is formulated as,
\begin{equation}
    \hat\vx = \{\hat\vx^{(1)}, \hat\vx^{(2)}, \cdots, \hat\vx^{(V)}\}, \ \hat\vx^{(v)} \in \R^{d_v}
\end{equation}
\begin{equation}
    \hat\vx^{(v)} = g^{(v)}(\vz; \vtheta_D^{(v)})
\end{equation}

\subsubsection{Clustering Module}
\label{subsubsec: CM}
The auto-encoder we have designed extracts robust representations and captures the inherent structures of data. However, these inherent structures may not necessarily follow a cluster-oriented distribution. To enhance the clustering performance, we introduce a Clustering Module inspired by DEC \citep{DEC}. 
Below we describe its procedures. First, after auto-encoder pretraining, a traditional clustering method is adopted to initialize the cluster centers $\vtheta_C \in \R^{d_c \times d_e}$ from embedding $\vz$. Then, during each iteration of joint training, a similarity matrix $\vq \in [0,1]^{N \times d_c}$ is generated between $\vtheta_C$ and embedding $\vz$ with Student's t-distribution, in which $q_{ij}$ represents the possibility sample $i$ belongs to cluster $j$, and $\vc \in \R^{d_c}$ defined in equation \ref{equation: cluster} is a row of $\vq$. With the similarity matrix $\vq$, we calculate the target distribution $\vp$ as,
\begin{equation}
    p_{ij} = \frac{q_{ij}^2/f_j}{\sum_{j=1}^{d_c} (q_{ij}^2/f_j)}
\end{equation}
in which $f_j$ is the soft cluster size $\sum_{i=0}^{N-1} q_{ij}$. The training target is the KL-Divergence between $\vp$ and $\vq$ (equation \ref{equation: clu loss}) which is defined in Section \ref{subsubsec: Loss}. 
Finally, after joint training, the clustering result can be obtained by finding the maximum possibility in each row of $\vq$. The detailed training procedures are described in Section \ref{subsubsec: TS}.

\subsection{Data Augmentation and Imputation}
\label{subsec: DAI}
In this section, we introduce Data Augmentation and KNN Imputation ($KIDA(\cdot)$ in equation \ref{equation: KIDA}), which are two key data processing strategies of our approach for handling incomplete multi-view data. 

\subsubsection{KNN Imputation}
\label{subsubsec: KNN}
The KNN search is conducted separately in each view, taking into account the incomplete condition. For existing views, the KNN is directly obtained, while for missing views, existing views from the same sample are used for searching KNN. Specifically, for a missing view $\vx_i^{(v)}$ of sample $\vx_i$, we first find all existing views $\vx_i^{(b)}$ of the same sample. Then we iterate through the KNN samples $\vx_j^{(b)}$ of these existing views to check if $\vx_j^{(v)}$ exists. If it does, we appended it to the KNN list of the missing view. Finally, we select the top $k$ samples from the KNN list as the imputation for the missing view. If the length of the KNN list is less than $k$, then the remaining positions are filled with zeros. Along with the KNN imputation $\bar\vx$, a missing indicator matrix $\bar\vm \in \R^{k \times V}$ is generated, where 1 represents a position filled with a KNN sample and 0 represents a position filled with zeros. The detailed procedures are listed in Algorithm \ref{algorithm: KNN}, Appendix \textbf{\ref{App: Method}}.

\subsubsection{Data Augmentation}
Our framework is inspired by the denoising auto-encoder, which helps to learn robust representations by introducing noise during training. Three types of noise are designed, including Gaussian noise, random dropout, and view dropout. Gaussian noise helps prevent overfitting by introducing variability in the input data. Random dropout, functional as a regularization technique, encourages the model to learn more robust features by forcing it to rely on different subsets of the input data. View dropout is a noise specifically designed for the IMVC task. It randomly drops out one or more views from the input data during training, encouraging the model to learn representations that are more robust to view missing conditions. 

Below are the formulations of the three kinds of augmentation we used. By combining equation \ref{equation: VD} and \ref{equation: GNRD}, the KIDA operation in equation \ref{equation: KIDA} is obtained. 
\begin{equation}
\label{equation: VD}
    \bar\vx', \bar\vm' = KNNI(\vx, \vm \odot \vm_{\sss VD}, X, M), \ P(\vm_{\sss VD} = 0) = \phi_1 
\end{equation}
\begin{equation}
\label{equation: GNRD}
    \bar\vx' = (\bar\vx' + \phi_2 \vn) \odot \vm_{\sss RD}, \ \vn \sim \mathcal{N}(0, 1), \ P(\vm_{\sss RD} = 0) = \phi_3
\end{equation}
The view dropout augmentation is shown in equation \ref{equation: VD}, where random views are dropped out with a possibility of $\phi_1$, before the KNN Imputation $KNNI(\cdot)$ (Algorithm \ref{algorithm: KNN}) step. The dropout mask is applied with element-wise multiplication, denoted by $\odot$. The masked views are regarded as missing views in both the KNN Imputation and the auto-encoder network. After the KNN Imputation, in equation \ref{equation: GNRD}, Gaussian noise is added to the input data to introduce variability, whose intensity is controlled by hyperparameter $\phi_2$. After that, random values in the input are set to zero with a probability of $\phi_3$, which is the random dropout augmentation. Finally, the augmented input $\bar\vx'$ along with the corresponding missing indicator matrix $\bar\vm'$ can be used for training. 

\subsection{Training Strategy and Loss Function}
\label{subsec: TSLF}

\subsubsection{Loss Function}
\label{subsubsec: Loss}
In the training process, we utilize a combination of 3 loss functions formulated as follows:
\begin{equation}
\label{equation: Loss}
    L(\vx, \hat{\vx}', \vz, \vz', \vc) = L_{rec}(\vx, \hat{\vx}') + \lambda_1 L_{aug}(\vz, \vz') + \lambda_2 L_{clu}(\vc)
\end{equation}

$L_{rec}$ corresponds to the reconstruction loss of the auto-encoder for learning meaningful latent representations, formulated as:
\begin{equation}
\label{equation: rec loss}
    L_{rec}(\vx, \hat\vx') = \sum_{v=1}^{V}(||\hat\vx'^{(v)} - \vx^{(v)}||^2 \odot \vm_v)
\end{equation}
The missing indicator $\vm_v$ is element-wisely multiplied so only the mean square errors of existing views are calculated. Note that during training, the network output is $\hat\vx'$ as the input is the augmented input $\bar\vx'$ (equation \ref{equation: Encoder} and \ref{equation: Decoder}), so the network learns to reconstruct dropped out views with KNN hints and cross-view correlation and thus, learns to predict missing views' information implicitly. 

$L_{aug}$ is the embedding robustness loss, which encourages the learned representations to be consistent when augmentations are applied, promoting the robustness of the learned representations, formulated as:
\begin{equation}
\label{equation: aug loss}
    L_{aug}(\vz, \vz') = -log \frac{e^{-||\vz'_i-\vz_i||}}{\sum_{j=1}^{B}e^{-||\vz'_i-\vz_j||}}
\end{equation}
Though our embedding robustness target is equivalent to minimizing the distance between $\vz'$ and $\vz$ (equation \ref{equation: Encoder}), we design $L_{aug}$ based on cross-entropy loss within the training mini-batch. $B$ in the equation represents the training batch size. This design simultaneously minimizes the distance between $\vz'_i$ and $\vz_i$ and maximizes the distances between $\vz'_i$ and embeddings of other samples $\vz_j, j \neq i$, preventing the embedding space from collapsing.

$L_{clu}$ is the DEC-based clustering loss \cite{DEC} for the Clustering Module (Section \ref{subsubsec: CM}), which optimizes embeddings for clustering with gradients from high-confidence samples. 
\begin{equation}
\label{equation: clu loss}
    L_{clu}(\vc) = KL(\vp'||\vq') = \sum_{i=1}^B\sum_{j=1}^{d_c} p'_{ij} log \frac{p'_{ij}}{q'_{ij}}
\end{equation}
It is formulated as the KL-divergence between distribution $\vp'$ and $\vq'$ computed from augmented input $\vx'$ during training.

Hyperparameters $\lambda_1$ and $\lambda_2$ control the balance between different loss components. By jointly minimizing the three loss terms, our network can learn representations that are both informative and clustering-friendly.

\subsubsection{Training Strategy}
\label{subsubsec: TS}
The training process is divided into two stages. In the first stage, the auto-encoder is pre-trained using $L_{rec}$ and $L_{aug}$, focusing on learning robust representations.
Once the pre-training is complete, the Clustering Module is initialized. In the second stage, $L_{clu}$ is added for joint training to learn a clustering-friendly representation. 

The training process is controlled by 2 hyperparameters: $E_p$, which represents the number of pre-training epochs, and $E_j$, which represents the number of joint training epochs.
For a detailed description of the training process, please refer to Algorithm \ref{algorithm: training} in Appendix \textbf{\ref{App: Method}}.

\section{Experiments}
Please refer to Appendix \textbf{\ref{App: ID}} for the hyperparameter settings and design details in our experiments.

\begin{table}[hbp]
    \centering
    \caption{The statistic of 6 datasets used in our experiments.}
    \resizebox{1.0\linewidth}{!}{
    \begin{tabular}{cccccc}
    \toprule
        \textbf{Name} & \textbf{Views} & \textbf{Clusters} & \textbf{Samples} & \textbf{Dimensions} & \textbf{Type} \\
    \midrule
        Handwritten \citep{Handwritten} & 6 & 10 & 2000 & 240/76/216/47/64/6 & Image \\
        Caltech101-7 \citep{Caltech101-7} & 5 & 7 & 1400 & 40/254/1984/512/928 & Image \\
        ALOI\_Deep \citep{RecFormer} & 3 & 100 & 10800 & 2048/4096/2048 & Image \\
        Scene15 \citep{Scene15, Caltech101-7} & 2 & 15 & 4485 & 20/59 & Image \\
        BDGP \citep{BDGP, DSIMVC} & 2 & 5 & 2500 & 1750/79 & Image/Text \\
        Reuters \citep{Reuters, SURE} & 2 & 6 & 18758 & 10/10 & Text \\
    \bottomrule
    \end{tabular}
    }
    \label{table: dataset}
\end{table}

\begin{table*}[htp]
    \centering
    \caption{Comparison of our method with state-of-the-art approaches on 6 benchmark datasets. The results are averaged on missing rates $m_r=\{0,0.25,0.5,0.75\}$. The best result is highlighted in \textbf{bold} while the sub-optimal is \underline{underlined}. }
    \resizebox{0.9\linewidth}{!}{
    \begin{tabular}{c|ccc|ccc|ccc}
    \toprule
        Datasets & \multicolumn{3}{c|}{Handwritten} & \multicolumn{3}{c}{Caltech101-7} & \multicolumn{3}{c|}{ALOI\_Deep} \\
    \midrule
        Metrics & Acc(\%) & NMI(\%) & ARI(\%) & Acc(\%) & NMI(\%) & ARI(\%) & Acc(\%) & NMI(\%) & ARI(\%) \\
    \midrule
        Completer \citep{Completer} & $52.19 \pm 5.14$ & $54.67 \pm 3.60$ & $28.77 \pm 4.72$ & $62.89 \pm 8.02$ & $57.75 \pm 6.50$ & $41.05 \pm 9.33$ & $44.53 \pm 2.53$ & $75.45 \pm 1.18$ & $26.47 \pm 2.10$ \\
        DSIMVC \citep{DSIMVC} & $75.76 \pm 4.04$ & $71.32 \pm 2.77$ & $63.17 \pm 4.11$ & $70.06 \pm 3.95$ & $59.56 \pm 2.58$ & $52.06 \pm 3.52$ & $72.58 \pm 2.24$ & $91.21 \pm 0.67$ & $70.15 \pm 2.10$ \\
        SURE \citep{SURE} & $66.46 \pm 6.81$ & $61.74 \pm 4.59$ & $50.37 \pm 6.38$ & $66.97 \pm 5.94$ & $54.37 \pm 4.92$ & $46.86 \pm 6.71$ & $50.08 \pm 5.96$ & $86.39 \pm 1.83$ & $40.66 \pm 7.56$ \\
        DCP \citep{DCP} & $59.80 \pm 6.32$ & $62.73 \pm 3.92$ & $45.40 \pm 8.21$ & $55.26 \pm 10.75$ & $54.44 \pm 9.59$ & $40.75 \pm 13.03$ & $56.89 \pm 5.34$ & $86.82 \pm 2.01$ & $50.66 \pm 9.00$ \\
        CPSPAN \citep{CPSPAN} & $84.19 \pm 5.43$ & $81.49 \pm 2.65$ & $76.06 \pm 4.79$ & \underline{$82.07 \pm 4.41$} & \underline{$73.04 \pm 4.19$} & \underline{$68.55 \pm 5.73$} & $73.01 \pm 4.27$ & $92.40 \pm 1.07$ & $71.79 \pm 3.78$ \\
        RecFormer \citep{RecFormer} & \underline{$90.46 \pm 1.33$} & \underline{82.67 $\pm$ 0.99} & \underline{$80.12 \pm 1.81$} & $71.07 \pm 2.53$ & $63.87 \pm 2.49$ & $56.77 \pm 3.09$ & \underline{$86.57 \pm 1.54$} & \underline{$96.84 \pm 0.33$} & \underline{$86.01 \pm 1.55$} \\
        URRL-IMVC (ours) & \textbf{93.86 $\pm$ 2.51} & \textbf{89.90 $\pm$ 1.38} & \textbf{88.93 $\pm$ 2.58} & \textbf{93.19 $\pm$ 1.54} & \textbf{86.94 $\pm$ 1.46} & \textbf{86.45 $\pm$ 1.98} & \textbf{91.48 $\pm$ 2.21} & \textbf{97.50 $\pm$ 1.04} & \textbf{90.91 $\pm$ 2.70} \\
    \specialrule{0.8pt}{2.0pt}{2.0pt}
        Datasets & \multicolumn{3}{c}{Scene15} & \multicolumn{3}{c|}{BDGP} & \multicolumn{3}{c}{Reuters} \\
    \midrule
        Metrics & Acc(\%) & NMI(\%) & ARI(\%) & Acc(\%) & NMI(\%) & ARI(\%) & Acc(\%) & NMI(\%) & ARI(\%) \\
    \midrule
        Completer \citep{Completer} & \underline{$38.39 \pm 1.96$} & \textbf{42.09 $\pm$ 1.54} & \underline{$22.86 \pm 1.78$} & $54.91 \pm 5.99$ & $46.89 \pm 4.62$ & $22.20 \pm 6.30$ & $38.68 \pm 4.02$ & $22.04 \pm 4.44$ & $8.26 \pm 4.12$ \\
        DSIMVC \citep{DSIMVC} & $31.63 \pm 1.22$ & $35.50 \pm 0.74$ & 17.48 $\pm$ 0.67 & \textbf{94.63 $\pm$ 1.53} & \textbf{85.62 $\pm$ 2.29} & \textbf{87.53 $\pm$ 2.74} & $44.07 \pm 2.91$ & \textbf{33.27 $\pm$ 2.10} & \underline{23.69 $\pm$ 2.20} \\
        SURE \citep{SURE} & $37.83 \pm 1.83$ & $37.62 \pm 0.80$ & $21.03 \pm 0.93$ & $60.48 \pm 9.91$ & $40.41 \pm 9.37$ & $34.68 \pm 10.60$ & \underline{46.68 $\pm$ 3.63} & $26.26 \pm 3.32$ & $20.57 \pm 2.16$ \\
        DCP \citep{DCP} & 38.28 $\pm$ 1.63 & 41.69 $\pm$ 1.23 & 22.22 $\pm$ 1.70 & $50.98 \pm 5.90$ & $44.50 \pm 6.50$ & $18.67 \pm 6.87$ & $38.60 \pm 3.29$ & $21.79 \pm 4.84$ & $7.12 \pm 3.80$ \\
        CPSPAN \citep{CPSPAN} & $37.71 \pm 2.33$ & 41.38 $\pm$ 2.04 & $22.68 \pm 1.84$ & $76.93 \pm 9.26$ & $63.25 \pm 7.62$ & $59.91 \pm 10.32$ & $39.78 \pm 2.02$ & $14.55 \pm 2.07$ & $12.47 \pm 1.54$ \\
        RecFormer \citep{RecFormer} & $33.37 \pm 1.39$ & $35.31 \pm 0.94$ & $17.45 \pm 0.79$ & $51.89 \pm 2.92$ & $40.46 \pm 2.69$ & $19.52 \pm 2.33$ & $41.43 \pm 3.59$ & $18.38 \pm 2.25$ & $15.91 \pm 2.23$ \\
        URRL-IMVC (ours) & \textbf{41.18 $\pm$ 1.77} & \underline{41.87 $\pm$ 0.95} & \textbf{24.09 $\pm$ 1.09} & \underline{89.15 $\pm$ 5.07} & \underline{77.14 $\pm$ 6.25} & \underline{77.64 $\pm$ 8.03} & \textbf{48.63 $\pm$ 2.64} & \underline{28.94 $\pm$ 1.65} & \textbf{24.78 $\pm$ 2.59} \\
    \bottomrule
    \end{tabular}
    }
    \label{table: SOTA}
\end{table*}

\subsection{Datasets and Metrics}
Experiments were performed on 6 multi-view datasets varying in number of views and modal to validate the effectiveness of our method. The dataset characteristics are summarized in Table \ref{table: dataset}. We report the widely used metrics Clustering Accuracy (Acc), Normalized Mutual Information (NMI), and Adjusted Rand Index(ARI) as results. We run each experiment 10 times and report the average value and standard deviation (after $\pm$). Details about our experiment and view-missing settings can be found in Appendix \textbf{\ref{App: Datasets}}.

\subsection{Comparison with State-of-the-arts}
We compare our approach with several state-of-the-art DIMVC approaches listed in Table \ref{table: SOTA}. Other comparisons about different numbers of views, traditional IMVC methods, model parameters, and computational costs can be found in Appendix \textbf{\ref{App: Traditional SOTA}}.

\paragraph{Comparison on different datasets}
URRL-IMVC achieved state-of-the-art performance on the 6 benchmark datasets, surpassing most existing approaches, as indicated in Table \ref{table: SOTA}. Our approach consistently outperformed other SOTA methods across all evaluation metrics, except for the BDGP dataset and NMI on Scene15 and Reuters, where our approach is sub-optimal. This excellent clustering performance and stability can be attributed to our representation learning framework, which effectively captures the underlying data structure while remaining robust in the presence of missing views. Additionally, URRL-IMVC exhibited stability compared to other SOTA methods, with a relatively lower standard deviation across 10 experiments, thanks to the tailored components in the network to filter out noise. Notably, our approach excelled on datasets with more views, such as Handwritten and Caltech101-7. Together with experimental results in Table \ref{table: different views}, it showed that our framework successfully overcomes the drawbacks of cross-view contrastive learning. 
 
\begin{figure*}[htbp]
    \centering
    \subfloat[]{
        \includegraphics[width=0.32\linewidth]{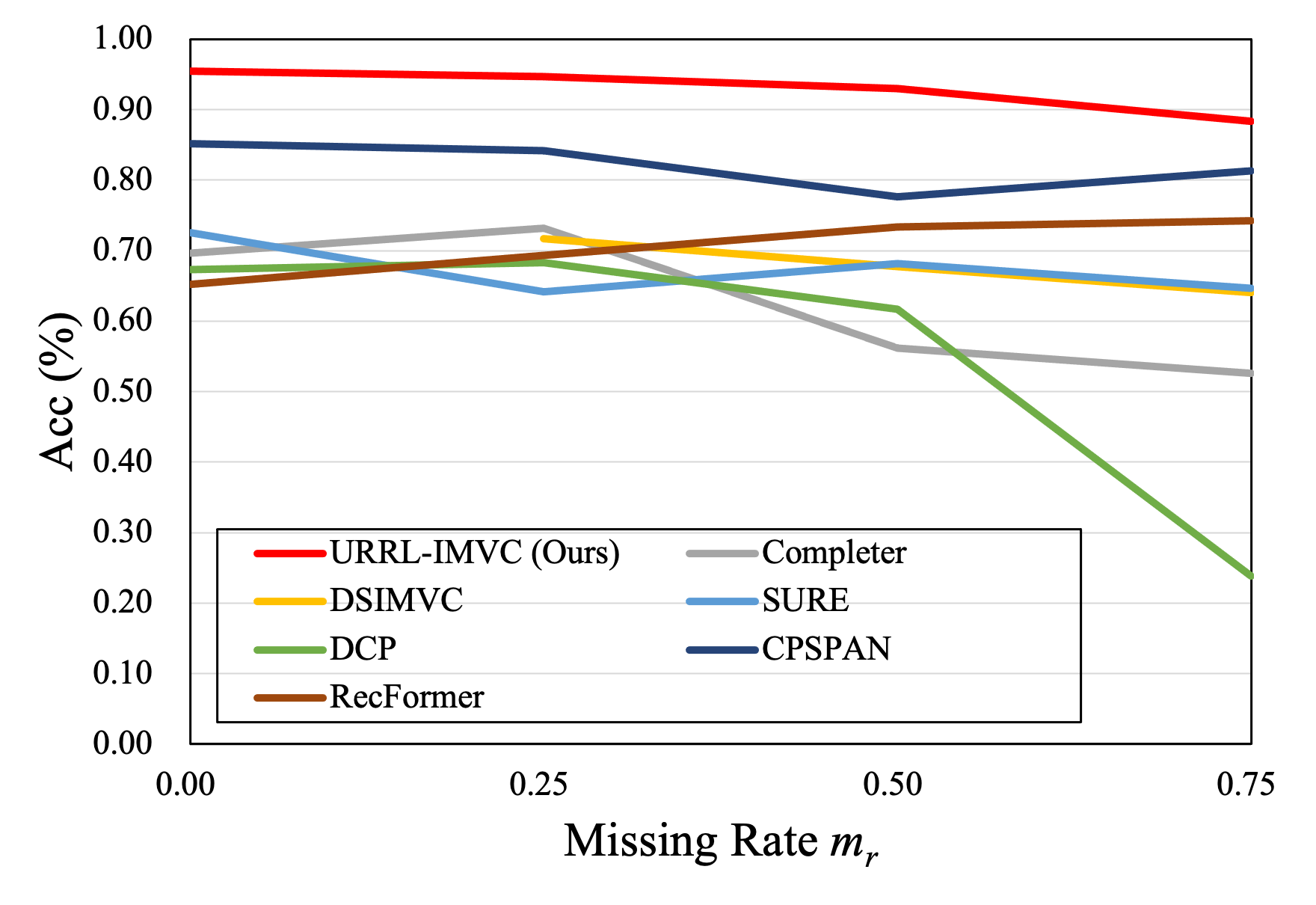}
        \label{fig: SOTA diff m_r 1}
    }
    \subfloat[]{
        \includegraphics[width=0.32\linewidth]{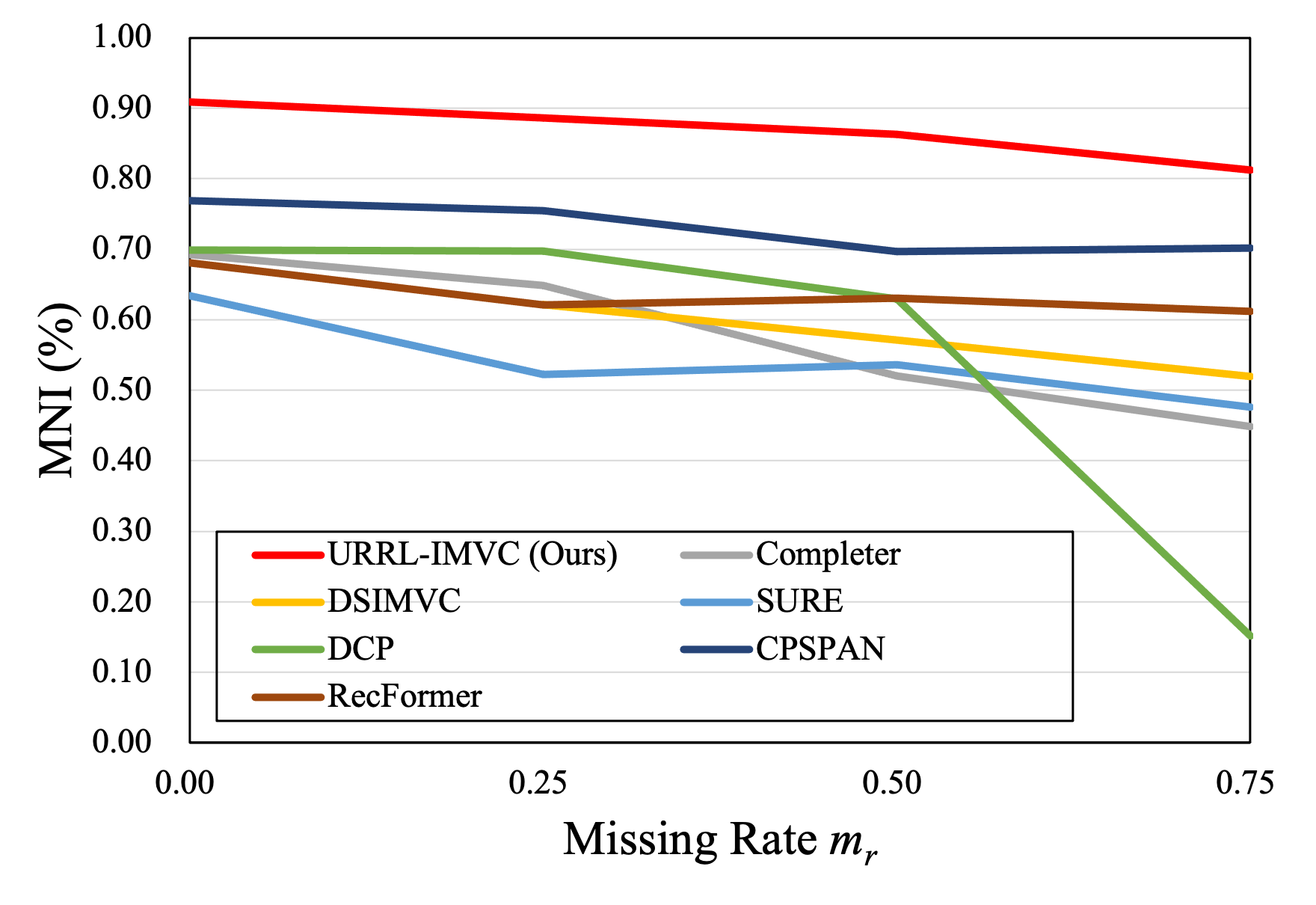}
        \label{fig: SOTA diff m_r 2}
    }
    \subfloat[]{
        \includegraphics[width=0.32\linewidth]{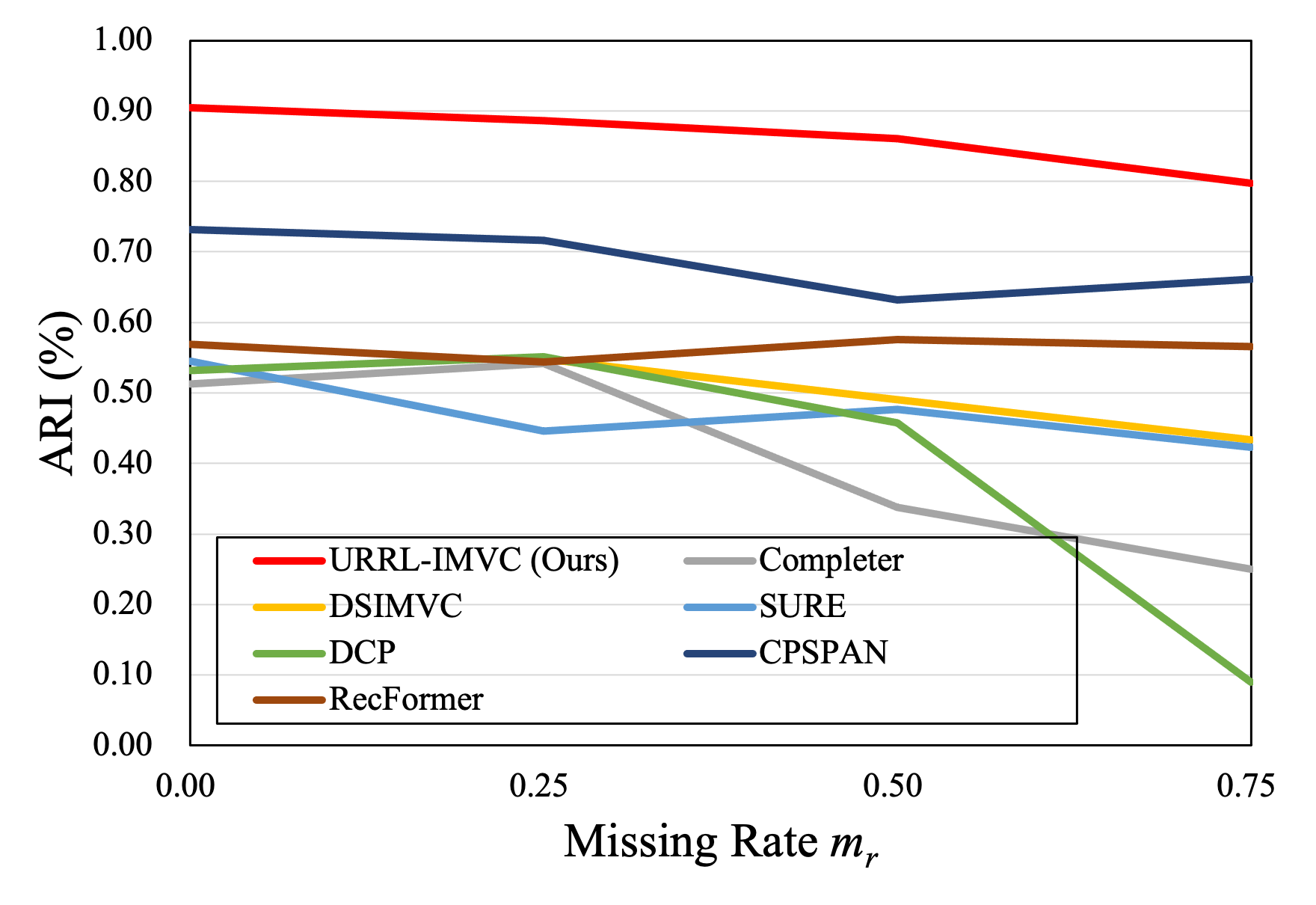}
        \label{fig: SOTA diff m_r 3}
    }
    \caption{Comparison with state-of-the-art approaches under different missing conditions on the Caltech101-7 dataset. The performance of each approach is reported using fold lines.}
    \Description{Our approach consistently outperforms other approaches under different missing rates.}
    \label{fig: SOTA diff m_r}
\end{figure*}

\paragraph{Comparison with different missing rates}
As depicted in Figure \ref{fig: SOTA diff m_r}, URRL-IMVC consistently outperformed other approaches, establishing an upper bound for clustering performance regardless of the missing rate ($m_r$). Our approach displayed better stability compared to other methods, with a gradual decrease in accuracy as the missing rate increased. In contrast, other approaches exhibited more fluctuation, rendering their results less predictable. Notably, DCP and Completer experienced a significant decline in performance when the missing rate reached 0.75, as they only trained their cross-view contrastive and recovery networks using complete samples. Insufficient training samples led to unsatisfactory recovery outcomes and fragile representations for clustering. In contrast, our approach focused on the robustness of the unified representation, allowing us to circumvent these limitations and achieve stable and high performance across varying missing rates. 

\subsection{Ablation Studies}
Unless otherwise specified, the experiments were conducted on the Caltech101-7 dataset with the missing rate $m_r=0.5$. In certain experiments, the Clustering Module was disabled to provide clearer observations of specific phenomena. Additional ablation studies regarding detailed designs and hyperparameters (e.g., output choice, $k$ in KNN Imputation, view dropout probability) can be found in Appendix \textbf{\ref{App: Ablation}}.

\subsubsection{Ablation on Modules} 
\begin{table}[htp]
    \centering
    \caption{Ablation study on our designed modules. We begin with the baseline model, which is a simple Transformer-based auto-encoder. Then different combinations of modules are incorporated to evaluate their contributions. ``KNNI'': KNN Imputation; ``Aug'': data augmentation and robustness loss; ``CDPE\&TAM'': CDPE and TAM described in NDE and VDE; ``CM'': clustering module and clustering loss.}
    \resizebox{\linewidth}{!}{
    \begin{tabular}{cccc|ccc}
    \toprule
        KNNI & Aug & CDPE\&TAM & CM & Acc(\%) & NMI(\%) & ARI(\%) \\
    \midrule
         &  &  &  & $76.93 \pm 4.35$ & 64.79 $\pm$ 2.18 & 56.52 $\pm$ 3.99 \\
    \midrule
        \checkmark &  &  &  & $83.68 \pm 2.79$ & 72.76 $\pm$ 2.21 & 68.60 $\pm$ 4.48 \\
         & \checkmark &  &  & 83.96 $\pm$ 3.11 & 74.39 $\pm$ 2.24 & 70.41 $\pm$ 3.63 \\
         &  &  & \checkmark & 85.60 $\pm$ 6.52 & 78.10 $\pm$ 4.46 & 75.29 $\pm$ 7.14 \\
    \midrule
        \checkmark & \checkmark &  &  & $84.77 \pm 3.59$ & 74.92 $\pm$ 2.54 & 71.76 $\pm$ 3.69 \\
        \checkmark &  & \checkmark &  & 82.05 $\pm$ 3.71 & 70.74 $\pm$ 3.44 & 63.92 $\pm$ 6.82 \\
        \checkmark &  &  & \checkmark & 89.95 $\pm$ 0.74 & 80.01 $\pm$ 0.96 & 79.55 $\pm$ 1.19 \\
        & \checkmark &  & \checkmark & 89.39 $\pm$ 3.12 & 81.74 $\pm$ 1.81 & 80.60 $\pm$ 2.76 \\
    \midrule
        \checkmark & \checkmark & \checkmark &  & $85.46 \pm 1.42$ & 75.25 $\pm$ 2.02 & 71.69 $\pm$ 2.44 \\
        \checkmark & \checkmark &  & \checkmark & 90.22 $\pm$ 5.20 & 83.21 $\pm$ 3.89 & 82.15 $\pm$ 5.67 \\
        \checkmark &  & \checkmark & \checkmark & \underline{91.90 $\pm$ 2.99} & \underline{84.29 $\pm$ 2.06} & \underline{83.76 $\pm$ 3.96} \\
    \midrule
        \checkmark & \checkmark & \checkmark & \checkmark & \textbf{93.35 $\pm$ 0.37} & \textbf{86.50 $\pm$ 0.61} & \textbf{86.25 $\pm$ 0.64} \\
    \bottomrule
    \end{tabular}
    }
    \label{table: Ablation Module}
\end{table}

In Table \ref{table: Ablation Module}, we present the results of our ablation study on the main modules we designed. First of all, our Unified auto-encoder framework sets a solid baseline. Then, our designed robustness strategies, KNN Imputation (KNNI), and Data Augmentation (Aug) significantly improve clustering performance and have approximately equal contributions. The Clustering Module (CM) also plays a vital role in some datasets, by learning clustering-friendly representations. However, directly applying it can result in unstable performance, as the DEC-based training is sensitive to initialization. While our tailored components, i.e., CDPE\&TAM help stabilize the learning. To summarize, the ablation study on modules reveals that the KNN Imputation, Augmentation, and Clustering Module are the three key components for improving clustering performance, while CDPE\&TAM is essential for stability. 

\subsubsection{Visualization} 

\begin{figure*}[htbp]
    \centering
    \subfloat[Raw data (38.43)]{
        \includegraphics[width=0.32\linewidth]{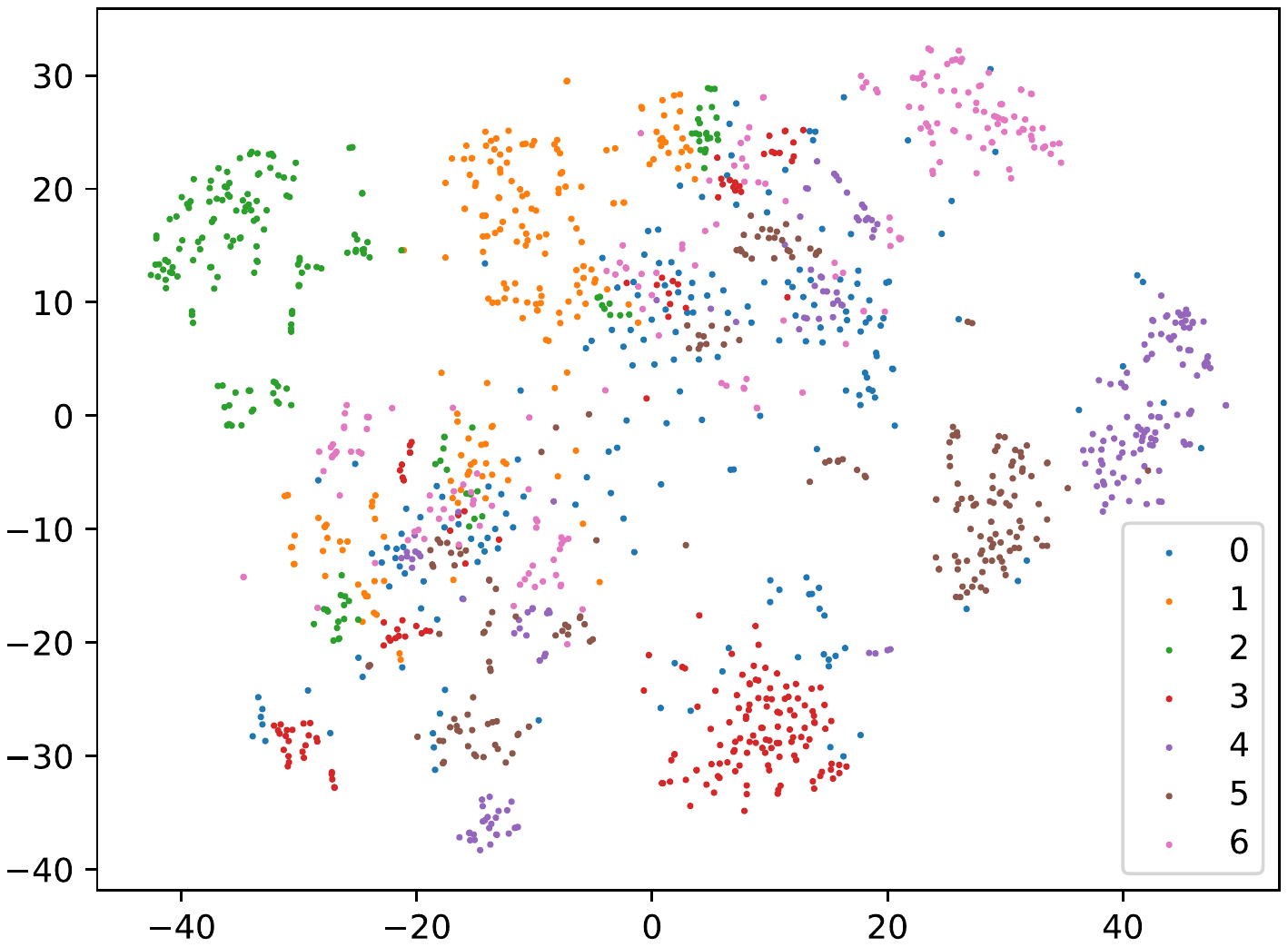}
        \label{fig: TSNE 1}
    }
    \subfloat[200 iteration (73.43)]{
        \includegraphics[width=0.32\linewidth]{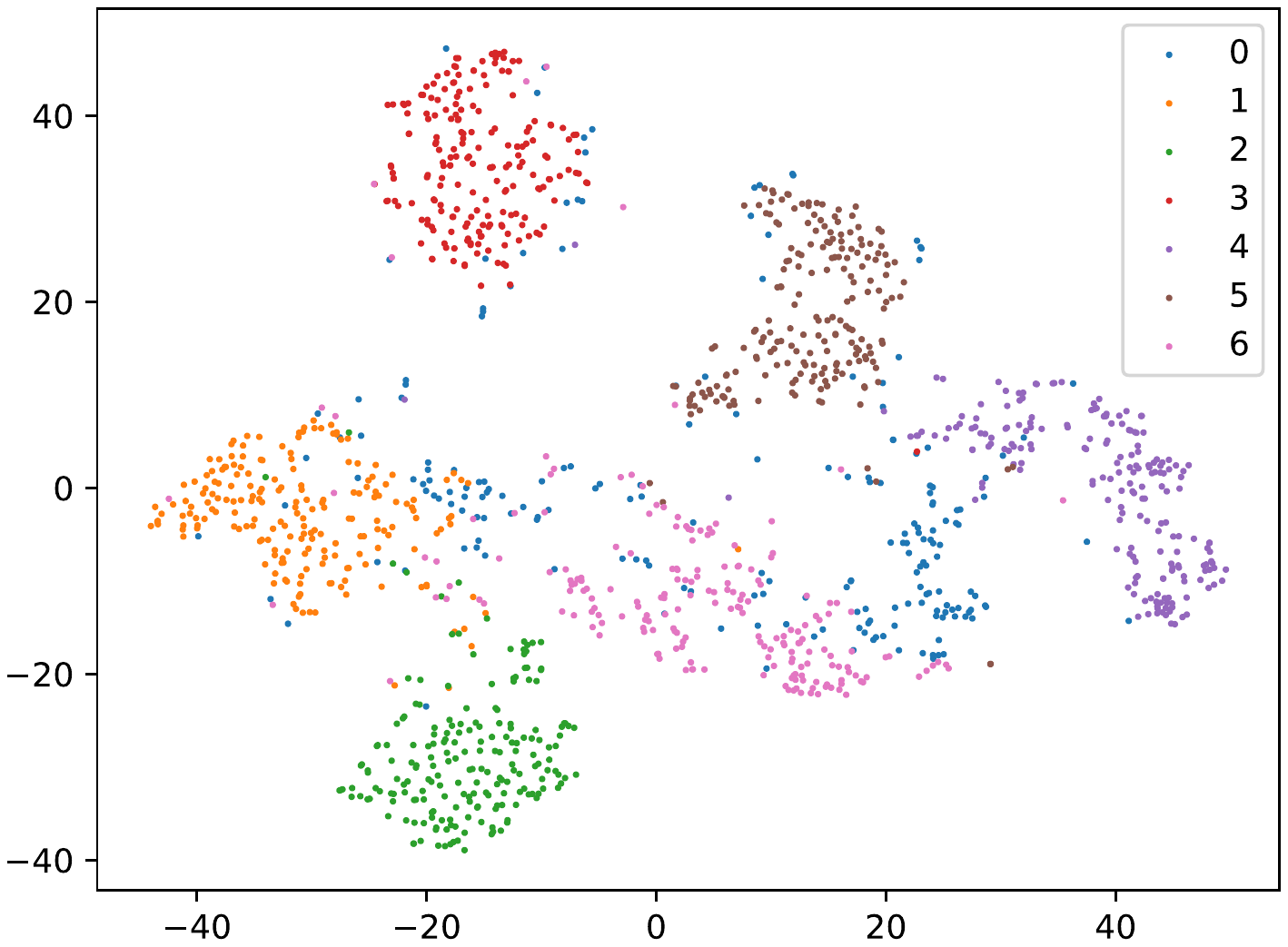}
        \label{fig: TSNE 2}
    }
    \subfloat[1600 iteration (89.86)]{
        \includegraphics[width=0.32\linewidth]{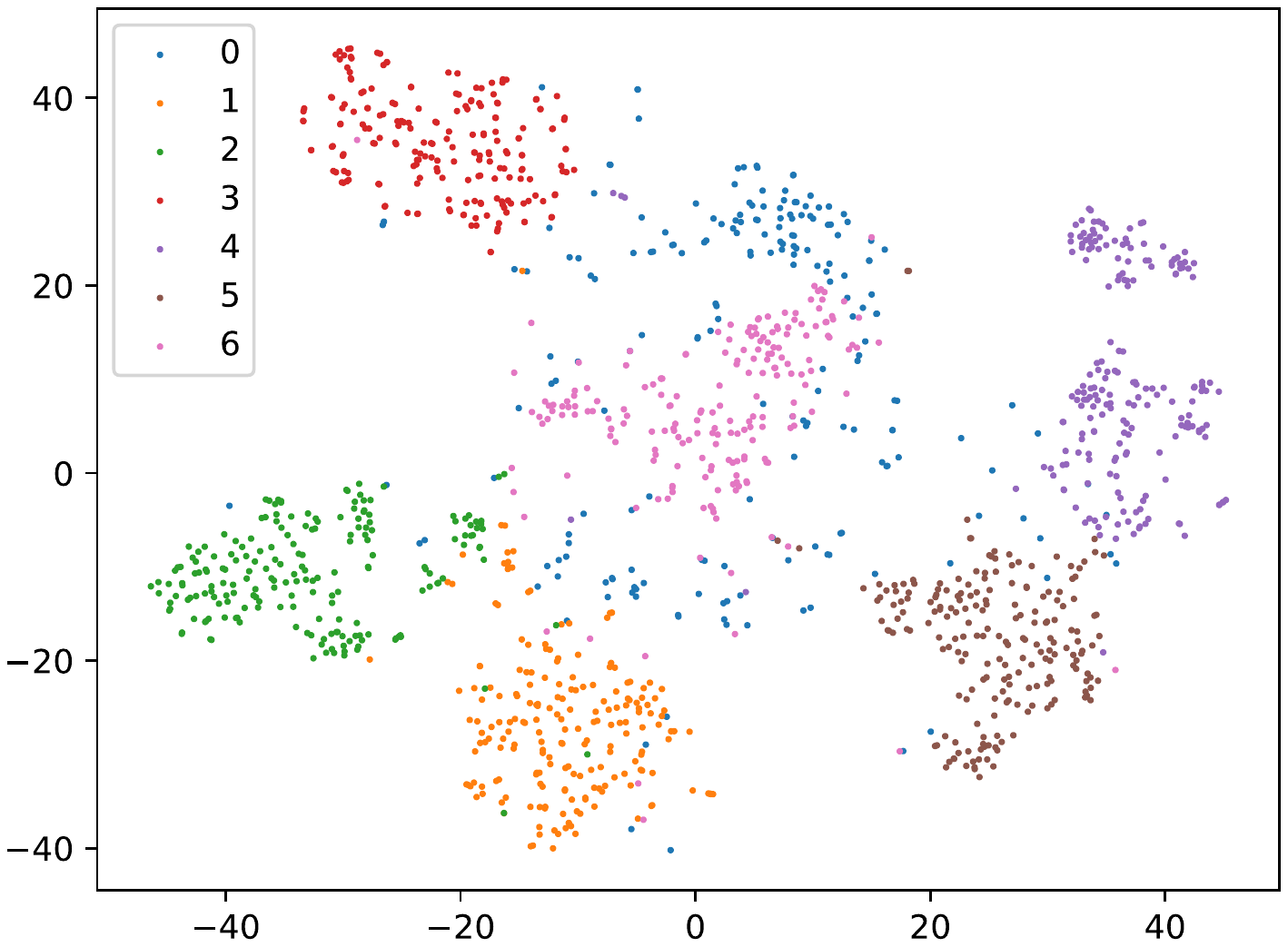}
        \label{fig: TSNE 3}
    } \\
    \subfloat[2200 iteration (87.86)]{
        \includegraphics[width=0.32\linewidth]{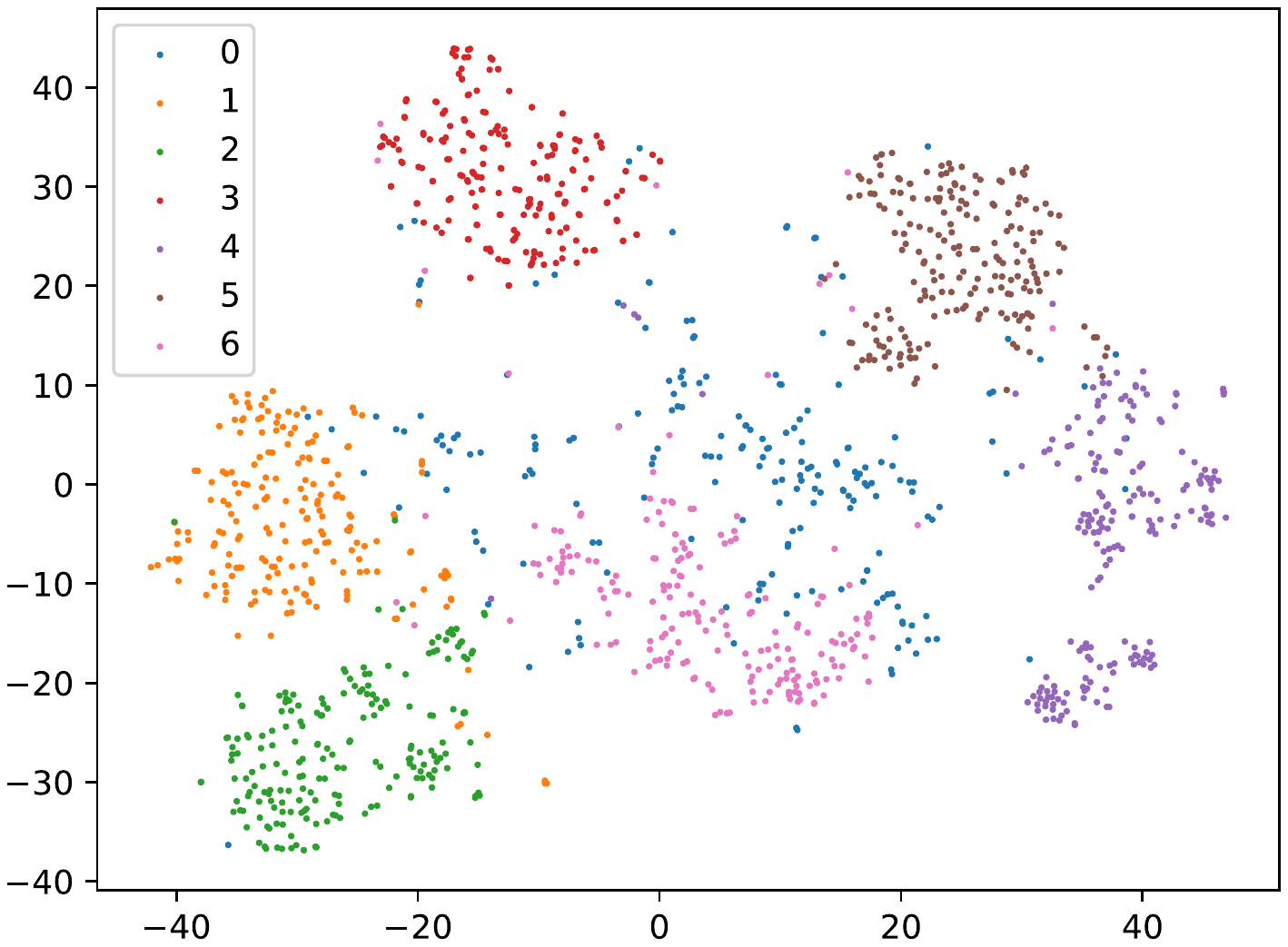}
        \label{fig: TSNE 4}
    }
    \subfloat[2400 iteration (90.14)]{
        \includegraphics[width=0.32\linewidth]{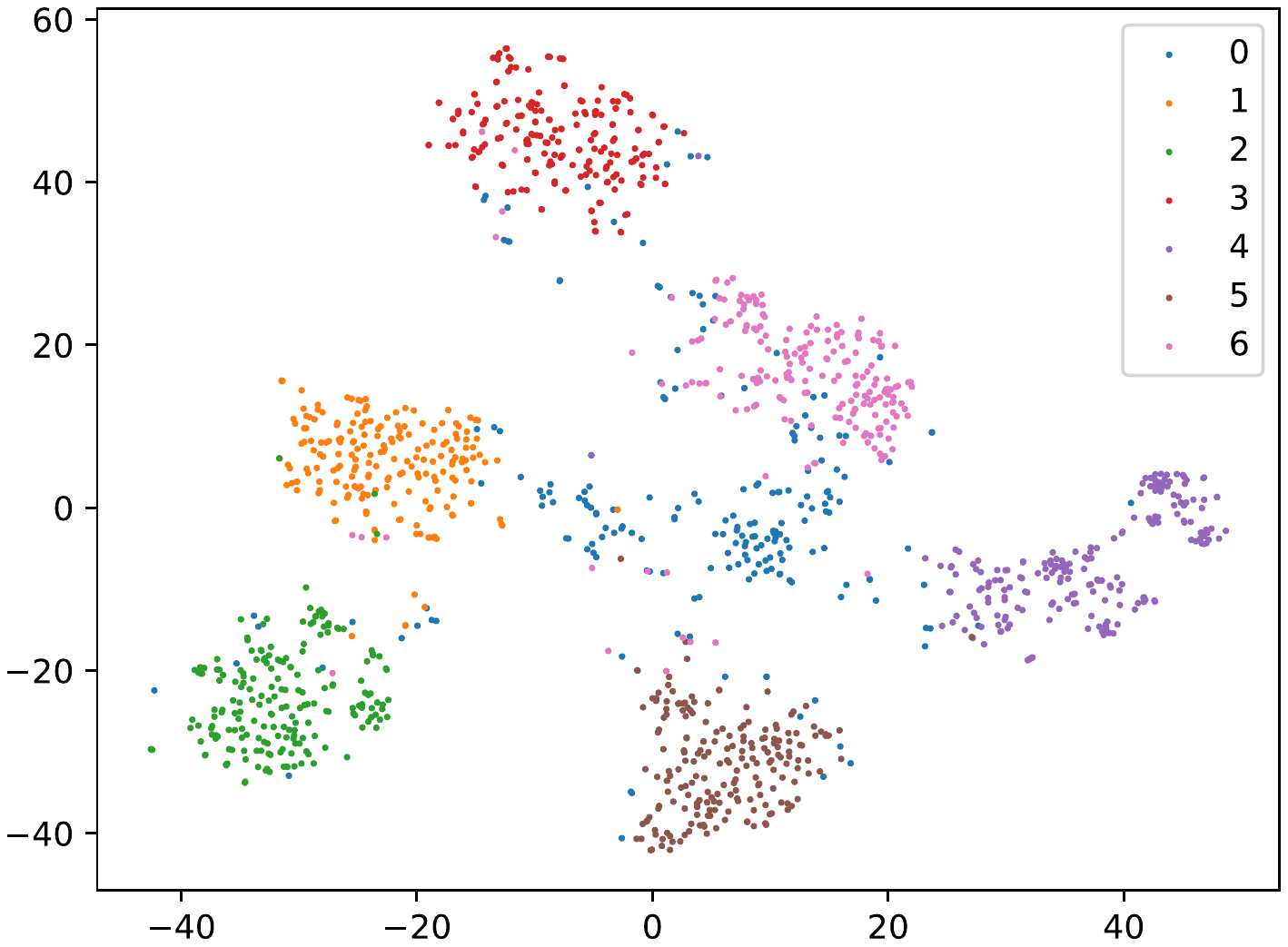}
        \label{fig: TSNE 5}
    }
    \subfloat[4400 iteration (94.14)]{
        \includegraphics[width=0.32\linewidth]{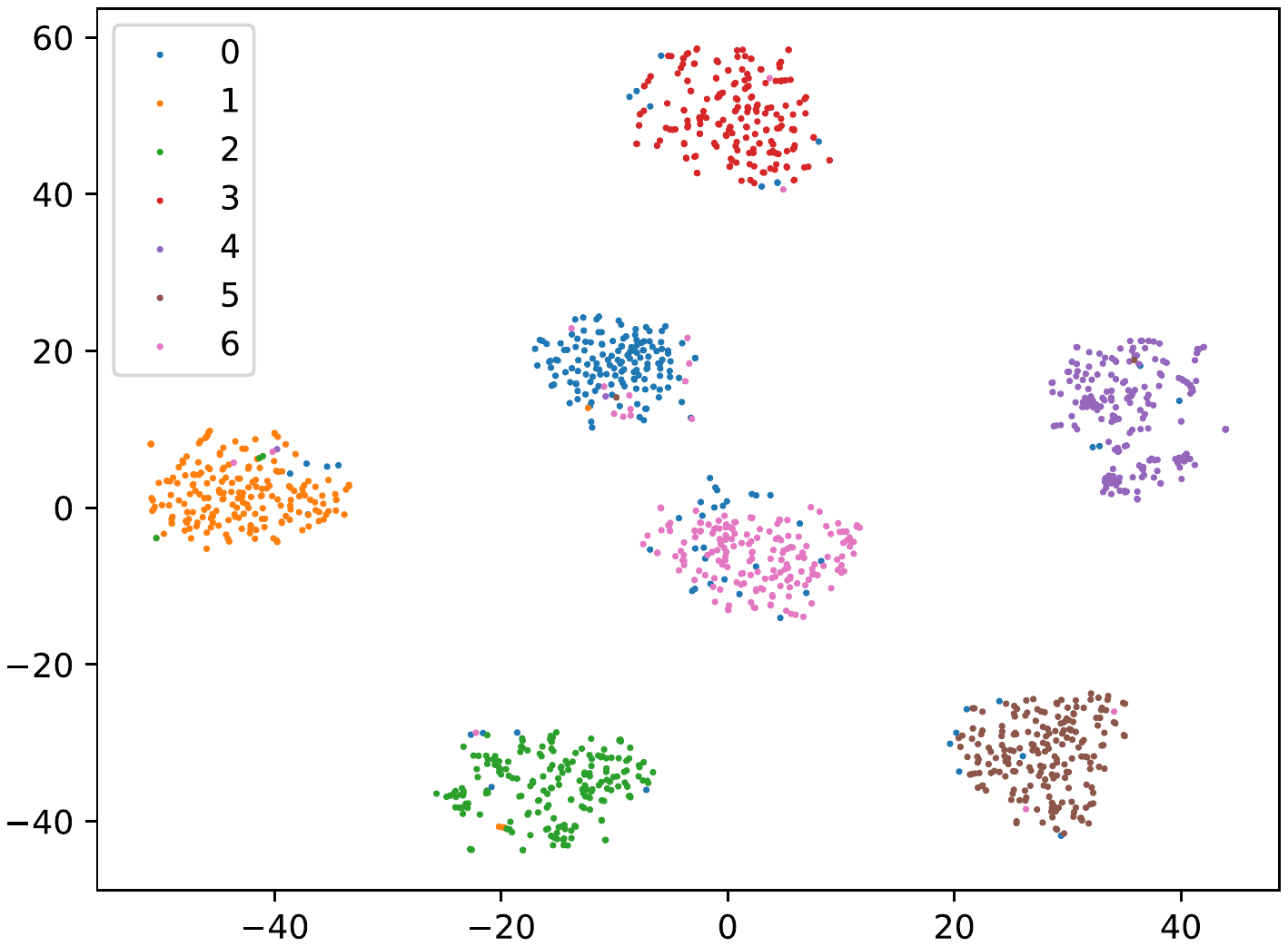}
        \label{fig: TSNE 6}
    }
    \caption{T-SNE visualization of the embeddings during the training process on the Caltech101-7 dataset. The iteration number and corresponding accuracy are recorded below each sub-figure. The training process consists of 4400 iterations, with the Clustering Module initialized at 2200 iterations.}
    \Description{The first half of iterations capture the inherent data structure and provide a good initialization for the Clustering Module.}
    \label{fig: TSNE}
\end{figure*}

Figure \ref{fig: TSNE} presents a T-SNE visualization of the embeddings during one training process. Initially in Figure \ref{fig: TSNE 1}, the multi-view raw data are concatenated as embeddings, and the visualization appears to be disorganized. After 200 iterations of training, in Figure \ref{fig: TSNE 2}, inherent structures start to be captured, and the pre-training peak accuracy (89.86) occurs at 1600 iterations \ref{fig: TSNE 3}. At 2200 iterations \ref{fig: TSNE 4}, the Clustering Module is initialized, and joint training with DEC-based clustering loss commences. 
The clusters become more compact after 200 iterations of joint training, as depicted in Figure \ref{fig: TSNE 5}. Finally, at the end of the training, as shown in Figure \ref{fig: TSNE 6}, the clusters become very compact, numerous samples initially incorrectly clustered with low confidence are now corrected, and the accuracy reaches 94.14. 

\section{Conclusion}
In this paper, we proposed URRL-IMVC, a novel unified and robust representation learning framework for the incomplete multi-view clustering task. 
By leveraging an attention-based auto-encoder framework, we successfully fuse the multi-view information into a unified embedding, offering a more comprehensive solution compared to potentially limiting cross-view contrastive learning. Through the utilization of KNN imputation and data augmentation strategies, we directly acquire robust embeddings that effectively handle the view-missing condition, eliminating the need for explicit missing view recovery and its associated computation and unreliability. Furthermore, incremental improvements, such as the Clustering Module and customization of the Encoder, enhance the clustering stability and performance, achieving state-of-the-art results. This improved robust and unified representation learning framework acts as a powerful tool for addressing the challenges of IMVC and provides valuable insights for future research in this domain.

\begin{acks}
    This work was supported by the Zhejiang Provincial Natural Science Foundation of China under Grant No. LDT23F01013F01, and by the Fundamental Research Funds for the Central Universities.
\end{acks}

\bibliographystyle{ACM-Reference-Format}
\bibliography{umrl-kdd}

\appendix

\section{Method Appendix}
\label{App: Method}

The detailed KNN Imputation algorithm is described in Algorithm \ref{algorithm: KNN}, and the training procedure is described in Algorithm \ref{algorithm: training}.

\begin{algorithm}[htp]
\caption{Procedure of KNN Imputation} \label{algorithm: KNN}
\raggedright
\textbf{Input:} Target $\vx_i^{(v)}$, which is the $v$th view of the $i$th sample, dataset $X$, missing indicator matrix $M$, hyperparameter $k$.
\begin{algorithmic}[1]
    \IF{$M_{iv}=1$} 
        \STATE \# The view exists
        \STATE Return $\vx_i^{(v)}$'s KNN 
    \ELSE 
        \STATE \# The view is missing
        \STATE $a=0$, create empty KNN list
        \WHILE{$a<k$} 
            \STATE \# Traverse $k$ neighbors
            \STATE $b=1$ 
            \WHILE{$b<=V$} 
                \STATE \# Traverse all views
                \IF{$b=v$ or $M_{ib}=0$} 
                    \STATE \# $b$th view of target sample $\vx_i$ is missing
                    \STATE pass
                \ELSE 
                    \STATE \# $b$th view of target sample $\vx_i$ exists
                    \STATE Find $a$th nearest neighbor of $\vx_i^{(b)}$, denoted as $\vx_j^{(b)}$
                    \IF{$M_{jv}=0$} 
                        \STATE \# $v$th view of neighbor $\vx_j$ is missing
                        \STATE pass
                    \ELSE 
                        \STATE \# $v$th view of neighbor $\vx_j$ exists
                        \STATE Add $\vx_j^{(v)}$ to KNN list
                    \ENDIF
                \ENDIF
                \STATE $b=b+1$
            \ENDWHILE
            \STATE $a=a+1$
        \ENDWHILE
        \IF{KNN list length $<k$}
            \STATE Pad KNN list with zeros to length $k$
        \ELSE
            \STATE Choose top $k$ from KNN list
        \ENDIF
        \STATE Return KNN list
    \ENDIF
\end{algorithmic}
\textbf{Output:} KNN Imputation $\bar\vx_i^{(v)}$
\end{algorithm}

\begin{algorithm}[htp]
\caption{Training process of URRL-IMVC} \label{algorithm: training}
\raggedright
\textbf{Input:} Dataset $X$, missing indicator matrix $M$, hyperparameters. 
\begin{algorithmic}[1]
    \STATE Initialize model parameters $\vtheta_E$, $\vtheta_D$, and $\vtheta_C$. Pre-compute KNN-search results. epoch = 0, iteration per epoch = $I_e$
    \WHILE{$epoch < E_p$} 
        \STATE \# Stage 1: Pre-training
        \STATE iteration = 0
        \WHILE{$iteration < I_e$}
            \STATE \textbf{Pre-process}: KNN Imputation and Data Augmentation by equation \ref{equation: VD} and \ref{equation: GNRD}, and obtain processed data $\bar\vx$, $\bar\vx'$ and processed mask $\bar\vm$, $\bar\vm'$.
            \STATE \textbf{Forward}: network forward by equation \ref{equation: Encoder}, \ref{equation: Decoder}, and obtain embedding $\vz'$, $\vz$ and reconstruction $\hat\vx'$
            \STATE \textbf{Loss}: Compute loss by equation \ref{equation: Loss}, in which $L_{clu}$ is ignored, i.e., $\lambda_2 = 0$.
            \STATE \textbf{Backward}: Loss backward and update model parameters $\vtheta_E$ and $\vtheta_D$.
            \STATE iteration = iteration + 1
        \ENDWHILE
        \STATE epoch = epoch + 1
    \ENDWHILE
    \STATE Initialize cluster centers $\vtheta_C$.
    \WHILE{$epoch < E_p + E_j$} 
        \STATE \# Stage 2: Joint Training
        \STATE iteration = 0
        \WHILE{$iteration < I_e$}
            \STATE \textbf{Pre-process}: KNN Imputation and Data Augmentation by equation \ref{equation: VD} and \ref{equation: GNRD}, and obtain processed data $\bar\vx$, $\bar\vx'$ and processed mask $\bar\vm$, $\bar\vm'$.
            \STATE \textbf{Forward}: network forward by equation \ref{equation: Encoder}, \ref{equation: Decoder}, and obtain embedding $\vz'$, $\vz$, reconstruction $\hat\vx'$, and clustering result $\vc'$
            \STATE \textbf{Loss}: Compute loss by equation \ref{equation: Loss}.
            \STATE \textbf{Backward}: Loss backward and update model parameters $\vtheta_E$, $\vtheta_D$, and $\vtheta_C$.
            \STATE iteration = iteration + 1
        \ENDWHILE
        \STATE epoch = epoch + 1
    \ENDWHILE  
\end{algorithmic}
\textbf{Output:} Model parameters $\vtheta_E$, $\vtheta_D$, $\vtheta_C$, final clustering result $\vc$
\end{algorithm}

\section{Experiments Appendix}

\subsection{Implementation Details}
\label{App: ID}
We set most of the hyperparameters empirically with grid search, and the same setting is used for all experiments if not specifically mentioned. $\gamma$ in equation \ref{equation: TLAM} is set to -10. The hyperparameter $k$ in KNN Imputation is set to 4. $\phi_2$ and $\phi_3$, the data augmentation hyperparameter in equation \ref{equation: GNRD}, are fixed at 0.05, while $\phi_1$ in equation \ref{equation: VD} which controls view dropout possibility is set to be growing with the actual missing rate of the dataset, defined as,
\begin{equation}
\label{equation: aug ratio}
    \phi_1 = \epsilon + (1 - \epsilon) \times (1 - \frac{\sum_{i=0}^N \sum_{j=0}^V M_{ij}}{N \times V})^2
\end{equation}
in which we set $\epsilon = 0.15$. The loss weight hyperparameters $\lambda_1$ and $\lambda_2$ are set to 0.001 and 0.1 respectively. 
The embedding dimension $d_e$ is set to 256. Batch size $B$ is set to 64 for both training and testing and the learning rate is fixed at 3e-4 throughout training. A small weight decay of 4e-5 is used for less over-fitting. The training epoch parameter $E_p$ (Section \ref{subsubsec: TS}) is set to 100, 100, 15, and 50 respectively for the four datasets in Table \ref{table: dataset} to maintain roughly the same training iteration. As for $E_j$, we found that training with DEC-based loss on some datasets (ALOI\_Deep and Scene15 in this paper) can diverge, possibly due to imbalanced cluster size. For these datasets, we simply skip the second stage's joint training, i.e., $E_j=0$, while for other datasets $E_j=E_p$.

PReLU \citep{PReLU} is used as the activation function in the VDE and the Decoder. Dropout is not used in any modules of our network. Agglomerative clustering with ``ward'' linkage is used to initialize cluster centers in the Clustering Module. 

\begin{table*}[htbp]
    \centering
    \caption{The ablation study on view dropout augmentation probability $\phi_1$ from equation \ref{equation: VD}. We use grid search to determine the best value range of $\phi_1$ under different missing rates $m_r$ and try to design a mapping function from the actual missing rate to the desired $\phi_1$. Generally speaking, a larger missing rate requires a larger view dropout probability for augmentation, the $\phi_1$ value from the designed mapping function, equation \ref{equation: aug ratio}, is listed in the last column of the table.}
    \resizebox{0.6\linewidth}{!}{
    \begin{tabular}{c|ccccc|c}
    \toprule
        Parameter & $\phi_1 = 0$ & $\phi_1 = 0.15$ & $\phi_1 = 0.3$ & $\phi_1 = 0.45$ & $\phi_1 = 0.6$ & Equation \ref{equation: aug ratio} \\
    \midrule
        $m_r = 0.00$ & $89.30 \pm 1.86$ & \underline{$89.50 \pm 1.77$} & $88.91 \pm 1.89$ & \textbf{89.71 $\pm$ 1.19} & $86.94 \pm 2.12$ & $\phi_1=0.15$ \\
        $m_r = 0.25$ & $86.15 \pm 2.42$ & \underline{$88.12 \pm 1.90$} & \textbf{88.56 $\pm$ 1.55} & $86.51 \pm 3.33$ & $83.57 \pm 4.74$ & $\phi_1=0.17$ \\
        $m_r = 0.50$ & $83.35 \pm 1.94$ & \textbf{86.12 $\pm$ 1.57} & \underline{$85.61 \pm 2.08$} & $83.25 \pm 3.86$ & $82.25 \pm 4.82$ & $\phi_1=0.23$ \\
        $m_r = 0.75$ & $81.41 \pm 2.79$ & \underline{$81.78 \pm 3.86$} & \textbf{82.89 $\pm$ 4.51} & $80.82 \pm 3.58$ & $77.17 \pm 5.63$ & $\phi_1=0.32$ \\
        $m_r = 1.00$ & $76.31 \pm 3.00$ & $77.53 \pm 3.63$ & $77.02 \pm 3.91$ & \textbf{80.21 $\pm$ 2.94} & \underline{$79.19 \pm 3.67$} & $\phi_1=0.46$ \\
    \bottomrule
    \end{tabular}
    }
    \label{table: Augmentation}
\end{table*}

\begin{table}[htbp]
    \centering
    \caption{Comparison between our approach and cross-view contrastive learning-based approach (CPSPAN) on Caltech101-7 dataset with different numbers of views. The best results of CPSPAN are achieved with 4 views, while with 5 views for our approach.}
    \resizebox{\linewidth}{!}{
    \begin{tabular}{c|ccc|ccc}
    \toprule
        Views & \multicolumn{3}{c|}{CPSPAN \citep{CPSPAN}} & \multicolumn{3}{c}{URRL-IMVC (ours)} \\
    \midrule
         & Acc(\%) & NMI(\%) & ARI(\%) & Acc(\%) & NMI(\%) & ARI(\%) \\
    \midrule
        2 & $50.88 \pm 1.87$ & $45.27 \pm 2.50$ & $35.79 \pm 2.25$ & $58.36 \pm 3.01$ & $47.16 \pm 2.50$ & $39.40 \pm 2.71$ \\
        3 & 73.17 $\pm$ 4.27 & 61.40 $\pm$ 4.29 & 55.37 $\pm$ 5.46 & $77.60 \pm 0.88$ & 67.61 $\pm$ 0.98 & 63.97 $\pm$ 1.33 \\
        4 & \textbf{84.89 $\pm$ 2.15} & \textbf{75.37 $\pm$ 2.45} & \textbf{71.79 $\pm$ 3.26} & \underline{91.73 $\pm$ 0.47} & \underline{83.57 $\pm$ 0.68} & \underline{83.26 $\pm$ 0.76} \\
        5 & \underline{77.62 $\pm$ 4.74} & \underline{69.70 $\pm$ 4.04} & \underline{63.23 $\pm$ 5.51} & \textbf{92.95 $\pm$ 2.60} & \textbf{86.29 $\pm$ 1.76} & \textbf{86.02 $\pm$ 2.91} \\
    \bottomrule
    \end{tabular}
    }
    \label{table: different views}
\end{table}

\subsection{Datasets and Experiments Setting}
\label{App: Datasets}
Our chosen datasets vary in views (2–6), clusters (7–100), samples (1400–18758), modal (image/text), and feature types (deep/hand-crafted), providing a comprehensive evaluation of approaches. Two parameters missing number $m_n$ and missing rate $m_r$ are defined to control the missing conditions. We first select $N \times m_r$ samples as incomplete samples, then randomly select $m_n$ views of each incomplete sample as missing views. We fix $m_n$ and vary $m_r$ in our experiments, $m_n$ are fixed at 4, 3, 2, 1, 1, 1 for the 6 datasets respectively. Importantly, it should be noted that within the same set of experiments, we ensured that the input data and missing indicator matrix remained consistent across different methods, ensuring fair comparisons. For comparison with state-of-the-art methods in Table \ref{table: SOTA} we reproduce the results with their published code. Several prior works are difficult to adapt to different numbers of views, which could hinder real applications. We randomly select views when the dataset has more views than the model requires. 

\subsection{Comparison with State-of-the-art Methods}
\label{App: Traditional SOTA}

\subsubsection{Comparison with a different number of views} 
As we mentioned in the introduction, the effectiveness of the cross-view contrastive learning strategy diminishes due to less overlapped information between views. Observing from Table \ref{table: different views}, adding more views may harm the clustering performance of the cross-view contrastive learning-based approach, proving this point of view, and also being consistent with the theoretical analysis from \citep{ESSCA}. On the other hand, our approach stably benefits from more views in the dataset, overcoming this drawback. 

\subsection{Ablation Studies} 
\label{App: Ablation}

\begin{table}[htbp]
    \centering
    \caption{Ablation test on the output choice of VDE and NDE. "Mean" represents using the average of all output vectors from the Transformer as output. The $1^{st}$ represents using the first output vector, $2^{nd}$ represents the second output vector, and so on. "Concat+Linear" represents first concatenating the output vectors and then using a linear layer to map the new vector to the desired dimension.}
    \resizebox{0.6\linewidth}{!}{
    \begin{tabular}{c|cc}
    \toprule
        Output/Module & NDE & VDE \\
    \midrule
        $1^{st}$ & \textbf{93.36$\pm$0.89} & 89.73$\pm$3.26 \\
        $2^{nd}$ & 91.87$\pm$1.01 & 88.75$\pm$3.56 \\
        $3^{rd}$ & 90.47$\pm$3.15 & 87.23$\pm$3.67 \\
        $4^{th}$ & 89.36$\pm$4.33 & 89.94$\pm$4.15 \\
        $5^{th}$ & - & 87.14$\pm$3.88 \\
        Mean & 90.09$\pm$3.61 & \textbf{93.36$\pm$0.98} \\
        Concat+Linear & 83.23$\pm$7.16 & 91.13$\pm$3.38 \\
    \bottomrule
    \end{tabular}
    }
    \label{table: Output Choice}
\end{table}

\subsubsection{Ablation on Output Choice} 
We conducted the ablation test in Table \ref{table: Output Choice} to find the best output choice of both the Neighbor Dimensional Encoder (NDE, section \ref{subsubsec:NDE}) and the View Dimensional Encoder (VDE, section \ref{subsubsec:VDE}). It can be observed that choosing the first vector of the Transformer output sequence significantly outperforms other choices, and using the latter output vectors results in worse and worse performance. It is consistent with our point of view that NDE needs a bias on the most confident input (the center sample or the nearest neighbor), and further neighbors contain more noise to harm the final performance. For VDE the situation is different, using the average of all output vectors outperforms other choices, which is consistent with our point of view that VDE needs to be unbiased. Concatenation with linear layer does not perform well in both Encoders, possibly due to lack of supervision. 

\subsubsection{Ablation on View Dropout Probability $\phi_1$} 
We conducted a grid search to determine the best value range of view dropout augmentation probability $\phi_1$ under different missing rates $m_r$, and the results are shown in Table \ref{table: Augmentation}. For view complete condition, $\phi_1 < 0.6$ have similar performance, while for view incomplete condition, the desired $\phi_1$ ascends as the missing rate $m_r$ increases. According to this observation, we designed the mapping function in equation \ref{equation: aug ratio} to follow this ascending trend, and its value is listed in the last column of the table.

\begin{table}[htbp]
    \centering
    \caption{Ablation test on the hyperparameter $k$ in KNN imputation. The result is unimodal with the best $k=4$. Larger $k$ values tend to provide more stable results (smaller standard deviation).}
    \resizebox{\linewidth}{!}{
    \begin{tabular}{ccccc}
    \toprule
        $k=1$ & $k=2$ & $k=4$ & $k=8$ & $k=16$ \\
    \midrule
        $83.84 \pm 3.38$ & $84.84 \pm 2.82$ & \textbf{87.31 $\pm$ 2.01} & \underline{$87.00 \pm 2.46$} & $86.01 \pm 1.28$ \\
    \bottomrule
    \end{tabular}
    }
    \label{table: K}
\end{table}

\subsubsection{Ablation on hyperparameter $k$ for KNN} 
We conduct an ablation study on $k$ in KNN Imputation (\ref{subsubsec: KNN}) to examine its effect. A large increment can be observed comparing $k=4$ with $k=1$. However, the performance starts to drop as $k>4$, which we infer can be caused by the noise brought by further neighbors. On the other hand, larger $k$ also seems to benefit the stability of clustering.

\end{document}